\documentclass{article}

\usepackage{microtype}
\usepackage{graphicx}
\usepackage{subcaption}
\usepackage{booktabs} 

\usepackage{hyperref}

\usepackage{amsmath}
\usepackage[ruled,vlined]{algorithm2e}
\usepackage{float}
\usepackage{multirow}
\usepackage[table]{xcolor}

\DeclareMathOperator*{\argmin}{arg\,min}

\usepackage[preprint]{icml2026}

\usepackage{amsmath}
\usepackage{amssymb}
\usepackage{mathtools}
\usepackage{amsthm}

\usepackage[capitalize,noabbrev]{cleveref}

\theoremstyle{plain}

\theoremstyle{definition}

\theoremstyle{remark}

\usepackage[textsize=tiny]{todonotes}

\icmltitlerunning{OT on the Map: Quantifying Domain Shifts in Geographic Space}

\begin{document}

\twocolumn[
  \icmltitle{OT on the Map: Quantifying Domain Shifts in Geographic Space}



  \icmlsetsymbol{equal}{*}

    \begin{icmlauthorlist}
      \icmlauthor{Haoran Zhang}{equal,harvard}
      \icmlauthor{Livia Betti}{equal,boulder}
      \icmlauthor{Konstantin Klemmer}{legend,london}
      \icmlauthor{Esther Rolf}{boulder}
      \icmlauthor{David Alvarez-Melis}{harvard,kempner,msr}
    \end{icmlauthorlist}
    
    \icmlaffiliation{harvard}{Harvard University}
    \icmlaffiliation{kempner}{Kempner Institute}
    \icmlaffiliation{msr}{Microsoft Research}
    \icmlaffiliation{boulder}{University of Colorado Boulder}
    \icmlaffiliation{legend}{LGND AI, Inc.}
    \icmlaffiliation{london}{University College London}
    
    \icmlcorrespondingauthor{Haoran Zhang}{haoran\textunderscore zhang@fas.harvard.edu}
    \icmlcorrespondingauthor{Livia Betti}{livia.betti@colorado.edu}

  \icmlkeywords{Machine Learning, ICML}

  \vskip 0.3in
]



\printAffiliationsAndNotice{}  



\begin{abstract}
In computer vision and machine learning for geographic data, out-of-domain generalization is a pervasive challenge, arising from uneven global data coverage and distribution shifts across geographic regions. Though models are frequently trained in one region and deployed in another, there is no principled method for determining when this cross-region adaptation will be successful. A well-defined notion of distance between distributions can effectively quantify how different a new target domain is compared to the domains used for model training, which in turn could support model training and deployment decisions. In this paper, we propose a strategy for computing distances between \underline{geosp}atial domains that leverages geographic information with \underline{O}ptimal \underline{T}ransport methods (\textsc{GeoSpOT}). 
In our experiments, \textsc{GeoSpOT} distances emerge as effective predictors of cross-domain transfer difficulty. 
%
We further demonstrate that embeddings from pretrained location encoders provide information comparable to image/text embeddings, despite relying solely on longitude-latitude pairs as input. 
This allows users to get an approximation of out-of-domain performance for geospatial models, even when the exact downstream task is unknown, or no task-specific data is available. Building on these findings, we show that \textsc{GeoSpOT} distances can preemptively guide data selection and enable predictive tools to analyze regions where a model is likely to underperform. Our code is available at \url{https://github.com/haoranzhang7/GeoSpOT}.



\end{abstract}    

\section{Introduction}
\label{section:introduction}





Machine learning (ML) with geospatial data has been used for a range of impactful applications, such as crop yield prediction \citep{Ansarifar2021}, disaster forecasting \citep{Linardos2022}, and pollution monitoring \citep{Hu2017}. 
Due to the limited availability and quality of labeled datasets, domain scientists commonly train models on data-rich regions (often in Western countries) 
and then transfer the model's predictive capabilities to target data-poor regions (often in the Global South) via domain adaptation.
This approach is not straightforward. Distinct climates, cultures, and environmental conditions cause distribution shifts between regions \citep{Sudmanns2020,federici2021distributionshift}, 
and, as a result, ML models may struggle to generalize effectively to unseen geographic domains \citep{rolf2024missioncriticalsatellite}. 
Quantifying when and where a model can generalize has been a focus of recent work in geospatial domain adaptation \citep{meyer2021predicting}, which compares geospatial datasets through measures of similarity or distance. \looseness-1

When comparing geospatial datasets, two complementary notions of distance naturally arise. The first is based on geographic proximity; standard measures such as geodesic distance quantify how close two regions are in space. This approach is intuitive: geographically closer domains often share climatic, social, and environmental characteristics, which should enable more effective domain adaptation and knowledge transfer. \citet{tobler1970computer} formalized this intuition in the ``first law of geography'', which states that ``all things are related, but near things are more related.''
However, purely spatial distances capture only the physical separation between locations and ignore the \textit{data} collected at those locations---the features that predictive models actually operate on. A second notion of distance instead compares datasets in \textit{feature space}, for example by using Optimal Transport (OT) on feature embeddings of the data \citep{alvarez2020geometric}. Such feature-based distances directly reflect similarity in the observable data (e.g., images, text, or other measurements), but they require that the data already exist for both domains---a costly and often impractical assumption in real-world geospatial settings. Moreover, these distances are not inherently aware of geographic structure, leaving important information unused.

In this work, we combine these two perspectives by introducing \textsc{GeoSpOT}, a geographically-aware OT framework for comparing geospatial datasets. \textsc{GeoSpOT} integrates spatial and feature-based information by defining an OT ground cost that incorporates both feature similarity and geographic proximity. To incorporate location information in a semantically meaningful way, we leverage pretrained location encoders---models that take geographic coordinates (latitude, longitude) as input and produce embeddings that capture environmental and contextual characteristics of those locations. These encoders are pretrained in a self-supervised manner, often using CLIP-style image-location matching objectives \cite{klemmer2023satclip, geoclip}. As a result, \textsc{GeoSpOT} can be applied independent of any specific dataset or task, enabling distance computation even when only geographic coordinates for the regions of interest are available. \looseness-1

Our experiments show that \textsc{GeoSpOT} distances reliably estimate cross-domain transfer performance across diverse downstream tasks. These findings advance our understanding of geospatial domain transfer by showing that purely spatial distances are insufficient to explain transfer success. Beyond estimation, we illustrate the practical utility of \textsc{GeoSpOT} through data selection experiments, where training sets chosen using distance-aware criteria yield models with significantly improved generalization to new domains. \looseness-1

\section{Related Work}
\label{sec:related work}

\paragraph{Geographic distribution shifts.} 
Machine Learning models trained on geospatial data often face substantial performance degradation when deployed in new regions due to \textit{geographic distribution shifts}. Such shifts arise from regional variations in terrain, seasonality, and acquisition conditions \citep{ekim2025distribution, rolf2024missioncriticalsatellite}, as well as differences in sensing instruments and collection protocols \citep{lynch2021leveraging}. For instance, while satellite images of houses in the United States and Egypt depict the same conceptual object, differences in  architectural style, urban planning, and environmental factors produce distinct visual signatures. These shifts are exacerbated by the common practice of training on data-rich regions and deploying in under-sampled areas. Previous work provides empirical evidence for these distribution shifts, e.g., highlighting performance gaps when training data lack geographic diversity \citep{nachmany2019detecting, shankar2017no}. \looseness-1

Efforts to address such shifts in geospatial ML have focused primarily on \textit{post hoc} adaptation and robustness. This includes work on out-of-distribution detection \citep{gawlikowski2021out, dimitric2023nearest, le2024detecting, ekim2025distribution} and improved domain adaptation methods for remote sensing imagery \citep{lin2019spatially, fang2022confident, makkar2022adv, ismael2023unsupervised}. Recent studies have shown that incorporating relevant metadata---particularly geographic coordinates---into adaptation frameworks can enhance cross-region transfer \citep{yao2023improving, crasto2025robustness}. In contrast, our work focuses on \textit{predicting} domain transfer success \textit{a priori}, even for the challenging case where no data beyond coordinates is available for the domain of interest.  Previous work such as \citet{meyer2021predicting} estimate model applicability by quantifying feature-space distances for individual samples. We instead propose to compare entire geospatial data distributions, introducing a principled way to anticipate geospatial adaptation challenges before model training.

\paragraph{Dataset distances for domain adaptation.} Rigorous, well-defined methods for quantitatively comparing datasets can guide domain adaptation decisions. Several distinct notions of dataset distances have been explored to predict cross-domain transfer success. Discrepancy distances measure the similarity over a set of functions from a hypothesis class \cite{kifer2004detecting, ben2006analysis} yet are often computationally intractable or infeasible to compute. Other measures of distance employ information-theoretic measures of similarity \citep{achille2019task2vec} or learned domain relations \citep{yao2023improving}, but depend on the model used to determine distances. More recent work leverages Optimal Transport (OT) as a flexible way to define dataset distances, encoding aspects of the data that are most relevant for learning, with extensions incorporating label information \citep{courty2017joint, alvarez2020geometric, tan2021otce}, group structure \citep{redko2020co}, or hierarchical relationships \citep{yurochkin2019hierarchical}. Among these, \citet{tan2021otce} combine the OT distance between domains with a label-dependent task distance, but in our work with label-agnostic setting, this reduces to the standard OT (Sinkhorn) distance between domains. Most closely related to our work, \citet{alvarez2020geometric} incorporate label information by modifying the ground cost to combine feature- and label-based distances; our approach follows this hybrid-metric principle, replacing label distance with geographic distance. OT approaches have been applied to geospatial ML: namely, for target shift \citep{redko2019optimal} and training on noisy data \citep{damodaran2020entropic}. Our approach focuses on predicting cross-region transfer in the setting of geographically separate domains, explicitly encoding feature and spatial information in the OT ground cost. This geographically-aware OT formulation tailors dataset distances to geospatial applications, improving predictive power for transfer across regions.

\section{Geospatial Dataset Distances}
\label{sec:geographic optimal transport}

\begin{figure*}[t] 
\centering
\begin{subfigure}{.94\textwidth}
  \centering
  \includegraphics[trim={0 5cm 0 0},clip, width=\linewidth]{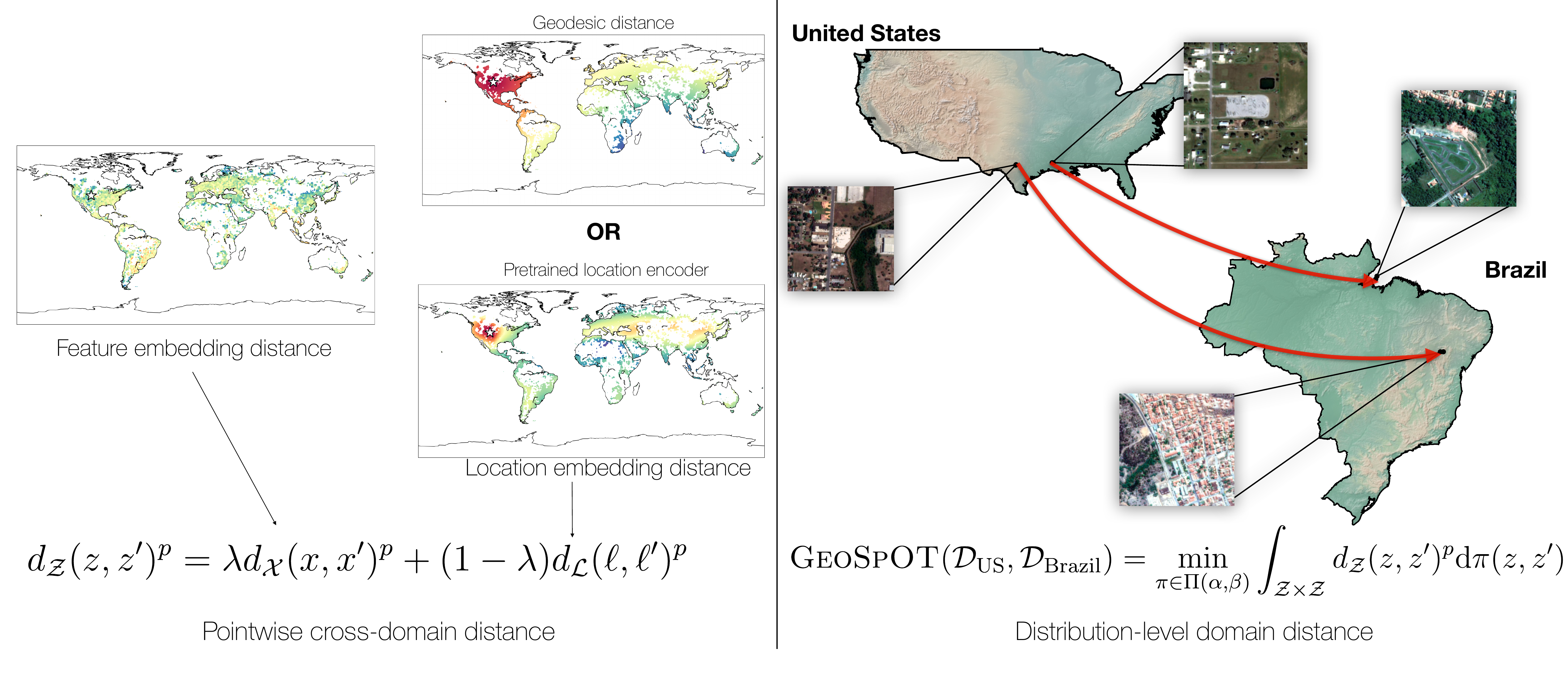}
\end{subfigure}%
\caption{\textbf{Overview of \textsc{GeoSpOT} domain distances:} (Left) Pointwise cross-domain computation according to feature (image) embedding distance, raw geographic distance, or location embedding distance, highlighting the difference between the approaches. (Right) Distribution-level domain distance between the United States and Brazil. Red arrows represent transported mass under the coupling resulting from solving the optimal transport problem. Coupled images  share similar visual features despite being far apart geographically.}
\label{fig:method_example}
\end{figure*}

\paragraph{Problem Formulation.}
Our goal is to define a notion of similarity or distance between \textit{geospatial domains} that jointly accounts for location and contextual information (e.g., images or text associated with each location). 
Given a source domain $\mathcal{D}_s$ and a target domain $\mathcal{D}_t$, we seek a principled distance between them that captures differences across both feature and location profiles.

\paragraph{Optimal Transport.} 
 Let $\mathcal{Z} \triangleq \mathcal{X} \times \mathcal{L}$ represent the joint feature-location space. We consider two finite datasets, $\mathcal{D}s = \{z_s^{(i)}\}_{i=1}^{n_s}$ and $\mathcal{D}_t = \{z_t^{(j)}\}_{j=1}^{n_t}$, corresponding to the source and target domains, which we treat as \textit{empirical probability measures}:
 $\alpha = \frac{1}{n_s} \sum_{i=1}^{n_s} \delta_{z_s^{(i)}}, \quad\beta = \frac{1}{n_t} \sum_{j=1}^{n_t} \delta_{z_t^{(j)}}$, where $\delta_z$ denotes the Dirac delta measure centered at $z$. 
 Given a pointwise ``ground'' cost function $c_\mathcal{Z}(\cdot, \cdot) : \mathcal{Z} \times \mathcal{Z} \rightarrow \mathbb{R}^{+}$ that quantifies the cost of matching elements across domains (and is often taken to be a distance metric $d_{\mathcal{Z}}$ in the ambient space), optimal transport lifts this into a notion of distribution-level discrepancy between the entire domains through the following optimization problem:
 \begin{equation*}
    \text{OT}(\alpha, \beta) = \min_{\pi \in \Pi(\alpha, \beta)} \int_{\mathcal{Z} \times \mathcal{Z}} c_\mathcal{Z}(z, z^\prime) \mathrm{d}\pi(z, z^\prime),
\end{equation*}
where $\Pi(\alpha, \beta)$ is the set of joint probability distributions over $\mathcal{Z} \times \mathcal{Z}$ with marginals $\alpha$ and $\beta$. Intuitively, OT finds a coupling $\pi$ that reallocates “mass” from points in the source domain to points in the target domain as efficiently as possible under the ground cost $c_\mathcal{Z}$. The value of the min-cost coupling between domains can be understood as a geometric-statistical quantification of their dissimilarity.



\paragraph{Geospatial Optimal Transport (\textsc{GeoSpOT}).}
Geospatial datasets contain both feature information (e.g., images or text) and the locations at which these features were collected, often stored as metadata. ``Geospatial domains'' can thus refer to distinct spatial units such as countries, continents, ecological regions, or grid cells on a longitude/latitude mesh. Their distinction may refer either to natural, physical attributes of the planet (e.g., ecological regions, climate zones) or refer to man-made definitions (e.g., voting districts, countries). Both carry different meanings, but can equally be characterized by distribution shifts among domains. To compute OT distances in a way that reflects both feature and location similarity, we first define a pointwise distance
on the joint space $\mathcal{Z} = \mathcal{X} \times \mathcal{L}$ as a convex combination of distances in each space: \looseness-1
\begin{equation}\label{eqn:ground dist all}
    d_\mathcal{Z}(z, z^\prime)^p = \lambda d_\mathcal{X}(x, x^\prime)^p + (1-\lambda) d_\mathcal{L}(\ell, \ell^\prime)^p
\end{equation}
The hyperparameter $\lambda$ controls the relative contribution of feature vs.~geographic distances.
For $d_\mathcal{X}$, we will often use a cosine-based dissimilarity between feature embeddings $x$ and $x^\prime$ extracted from pretrained image or text encoders.
For $d_\mathcal{L}$, we consider two alternatives: (1) raw geographic (arc) distances between latitude–longitude coordinates, and (2) cosine distance between pretrained location embeddings. 

When both $d_\mathcal{X}$ and $d_\mathcal{L}$ are valid metrics and $p \ge 1$, this $\ell_p$ combination defines a metric on the joint space $\mathcal{Z}$. Here, however, we will often take $d_\mathcal{X}$ to be a cosine-based dissimilarity, which does not strictly satisfy the triangle inequality. In such cases, $d_\mathcal{Z}$ should be interpreted more generally as a ground cost rather than a true metric—a distinction that does not affect the validity of the OT formulation, which only requires nonnegative costs.

Throughout our experiments,  we use the cosine distance $d_\mathcal{X}(x,x^\prime) = 1 - x^\top x^\prime / \|x\|\|x^\prime\|$ between feature embeddings extracted from pretrained image or text models. For geographic distance $d_\mathcal{L}$, we consider two variants: (1) the raw geographic (arc) distance $d_\text{arc}$ between the lat/lon coordinates, and (2) the cosine distance between pretrained location embeddings obtained from a location encoder. \looseness-1


With this, we define the Geospatial Optimal Transport (\textsc{GeoSpOT}) distance between domains as
\begin{equation}
    \textsc{GeoSpOT} (\mathcal{D}_s, \mathcal{D}_t) = \min_{\pi \in \Pi(\alpha, \beta)} \int_{\mathcal{Z} \times \mathcal{Z}} d_\mathcal{Z} (z, z^\prime)^p \mathrm{d}\pi(z, z^\prime).
\end{equation}


In practice, computing $d_\mathcal{Z}$ for \textsc{GeoSpOT} requires two design choices: (1) the relative weighting of feature vs. location information (via  $\lambda$ in \Cref{eqn:ground dist all}), and (2) the choice of geographic distance representation in computing $d_\mathcal{L}$. We explore both of these design choices in \Cref{sec: predict w dist}.

\section{Experimental Setup}
\label{section:experimental setup}

\subsection{Datasets}
\label{subsection:datasets}

We conduct our experiments on three datasets that differ in data modality, scale, and the nature of their class distributions and domain shifts. Data preprocessing details are provided in the Appendix and code release.

\vspace{-0.3cm}
\paragraph{Geo-YFCC.} The Geo-YFCC dataset \citep{dubey2021geoyfcc} is a large-scale domain generalization benchmark consisting of more than 1.1 million geotagged samples of 1,261 classes from 62 different country domains, subsampled from the YFCC100M collection of Flickr images \citep{thomee2016yfcc}. Each sample in the dataset has one or more associated labels. We convert this to single label classification by duplicating entries, consistent with the original work. The majority of samples also contain text information in the form of Flickr image description in the metadata. To explore robustness of \textsc{GeoSpOT} distances across visual and textual modalities, we use this dataset for two separate experimental conditions: one using the georeferenced image samples (Geo-YFCC-Image) and another using the georeferenced text captions (Geo-YFCC-{Text}).

\vspace{-0.2cm}
\paragraph{FMoW-Wilds.} FMoW-Wilds is a classification benchmark emphasizing distribution shifts across spatial and temporal dimensions \citep{wilds2021}, adapted from the original FMoW (Functional Map of the World) dataset \citep{fmow2018}. FMoW-Wilds contains RGB satellite images at size $224 \times 224$ pixels, each labeled with one of 62 land use or building classes. We focus on geospatial distribution shifts and disregard the temporal component. We further restrict to countries with abundant training data spanning most categories for reliable training and evaluation. This selection yields a subset of 152,196 samples from 25 countries around the globe.

\vspace{-0.2cm}
\paragraph{GeoDE.} GeoDE is a geographically diverse classification dataset with 61,940 crowd-sourced images, evenly distributed across 40 common object classes and 6 global regions \citep{ramaswamy2022geode}. To ensure sufficient data per domain for training, we keep only countries with $\geq$1,000 data samples, resulting in a subset of 61,378 images from 19 domains.

\subsection{Embeddings and Model Training}
\label{sec:embeddings_model_training}
We extract feature embeddings from image and text with commonly-used pretrained embedding models: a ResNet50 backbone pretrained on ImageNet \citep{he2015resnet}, and BERT, a transformer-based language model \citep{devlin-etal-2019-bert}. We generate location embeddings from longitude and latitude coordinates using the pretrained SatCLIP and GeoCLIP location encoders. SatCLIP embeds geographic coordinates with spherical harmonics and is pretrained via a contrastive learning objective pairing satellite imagery with coordinates \citep{klemmer2023satclip}.  GeoCLIP is another CLIP-based model that uses random Fourier features and hierarchical resolution to align locations with visual features from geo-tagged images \citep{geoclip}.

We compute \textsc{GeoSpOT} dataset distances  using the Python Optimal Transport (POT) library \citep{flamary2021pot, flamary2024pot}. To improve efficiency, we use the entropically regularized Sinkhorn divergence instead of solving the exact OT problem, with  regularization parameter $\epsilon= 0.01.$
For model training, we fine-tune pretrained ResNet50 encoders for image tasks and BERT models for text tasks, replacing the original classifier head with a linear layer sized to the target classes. Separate learning rates are used for the backbone and the new head, preserving pretrained 
representations while enabling faster adaptation in the task-specific layer. \looseness-1


\section{Predicting Domain Transfer Performance}
\label{sec: predict w dist}

We assess the quality of \textsc{GeoSpOT} distances by examining its correlation with out-of-distribution model performance of a trained classifier across different domain pairs. For each source-target domain pair $(\mathcal{D}_s, \mathcal{D}_t)$, we train two models $(\mathcal{M}_s, \mathcal{M}_t)$ on the respective training sets of $\mathcal{D}_s$ and $\mathcal{D}_t$, and we evaluate both models on the test set of the target domain. We measure the relative change ($\Delta$) in test performance with respect to in-distribution training as 
\begin{equation}\label{equation:test_acc_relative_change}
    \Delta_{\mathcal{D}_s, \mathcal{D}_t} = \frac{\text{Acc}_{\mathcal{D}_t}(\mathcal{M}_s) - \text{Acc}_{\mathcal{D}_t}(\mathcal{M}_t)}{\text{Acc}_{\mathcal{D}_t}(\mathcal{M}_t)} \times 100\%,
\end{equation}
where $\text{Acc}_{\mathcal{D}_t}(\mathcal{M}_s)$ represents performance of a model transferred from the source domain to the target domain, and  $\text{Acc}_{\mathcal{D}_t}(\mathcal{M}_t)$ represents in-sample performance of a model trained and tested on the target domain. This quantity, which is typically negative, can be interpreted as the penalty incurred when training on a source domain different from the target. Values closer to zero indicate better transferability, while larger negative values reflect poorer cross-domain generalization. 
Our goal is to estimate $\Delta_{\mathcal{D}_s, \mathcal{D}_t}$ with \textsc{GeoSpOT} distances, which depend only on the features and/or locations of the source and target domains (but not labels). To evaluate how well \textsc{GeoSpOT} distances correlate with domain transfer performance, we measure this association with Spearman's rank correlation ($\rho$), and the coefficient of determination, ($\mathcal{R}^2$) of \textsc{GeoSpOT} distances with $\Delta_{\mathcal{D}_s, \mathcal{D}_t}$ across pairs of source and target domains.


\begin{figure*}[!ht] 
\centering
\begin{subfigure}{\textwidth}
  \centering
  \includegraphics[width=\linewidth]{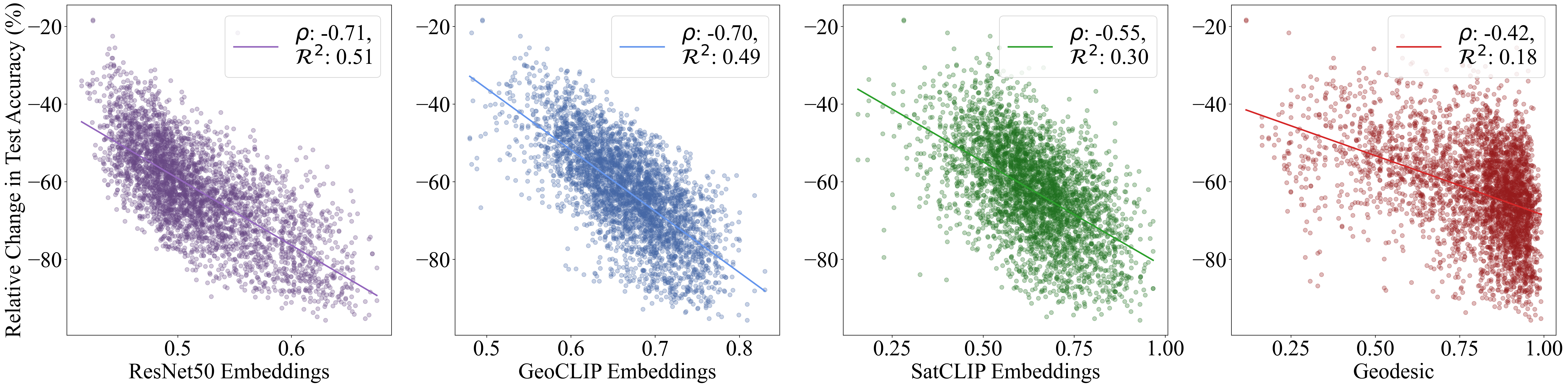}
  \caption{Geo-YFCC-Image}
  \label{subfig:trend_plots_geoyfcc_img}
\end{subfigure}

\medskip

\begin{subfigure}{\textwidth}
  \centering
  \includegraphics[width=\linewidth]{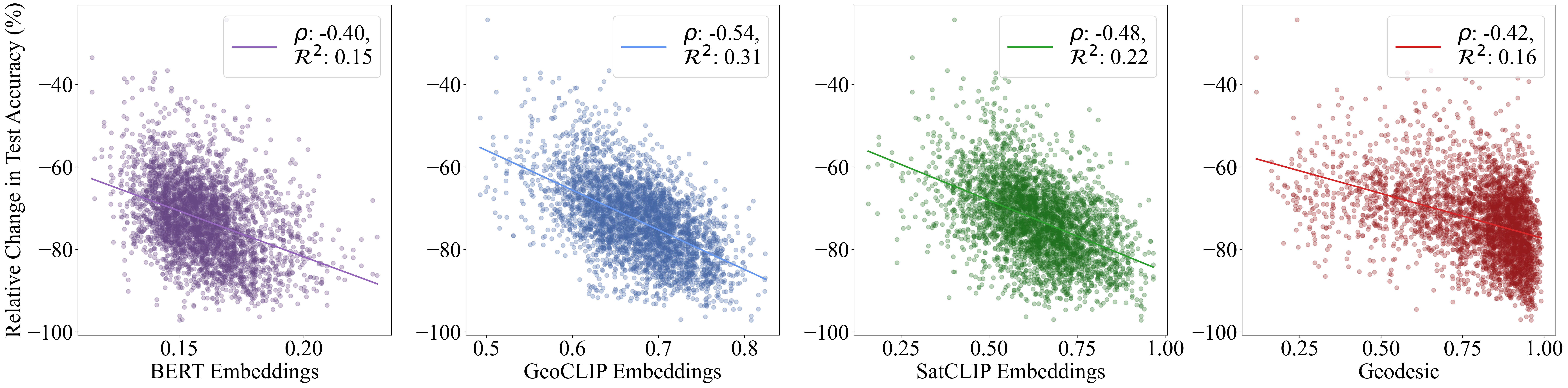}
  \caption{Geo-YFCC-Text}
  \label{subfig:trend_plots_geoyfcc_text}
\end{subfigure}

\medskip

\begin{subfigure}{\textwidth}
  \centering
  \includegraphics[width=\linewidth]{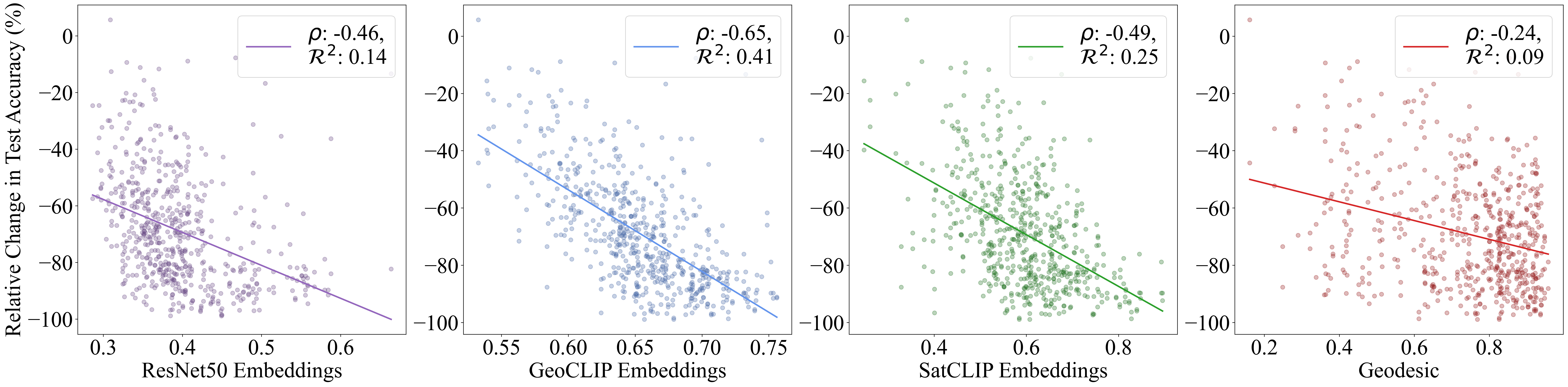}
  \caption{FMoW-Wilds}
  \label{subfig:trend_plots_fmow_wilds}
\end{subfigure}

\medskip

\begin{subfigure}{\textwidth}
  \centering
  \includegraphics[width=\linewidth]{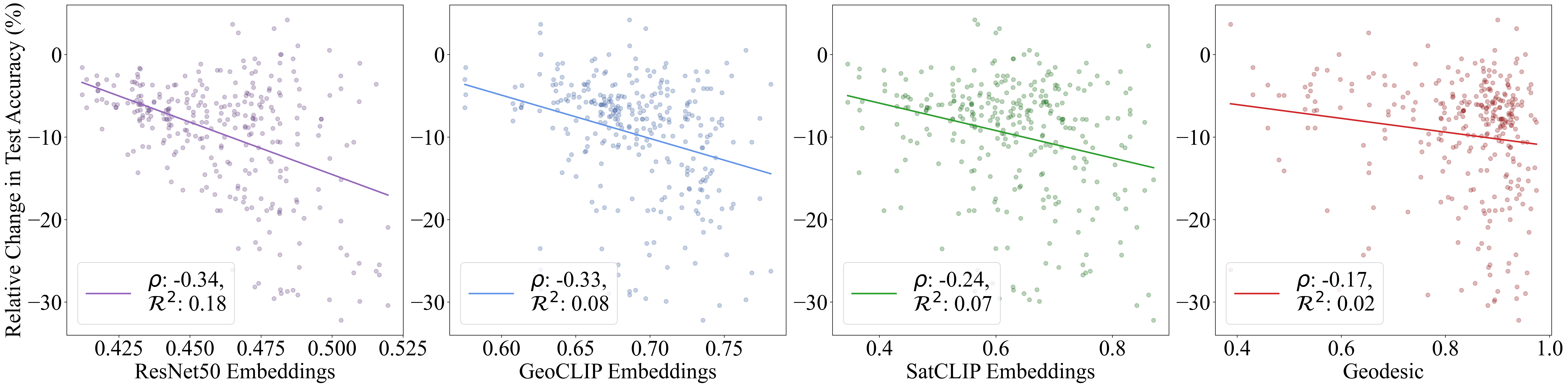}
  \caption{GeoDE}
  \label{subfig:trend_plots_geode}
\end{subfigure}

\caption{\textbf{\textsc{GeoSpOT} distances correlate with transfer performance (i.e., performance difference) between train and test domains}, measured in a zero-shot transfer setting. \textsc{GeoSpOT} distances are computed with a single modality ($\lambda =1$ for ResNet50/BERT, and $\lambda = 0$ for the location embeddings). Results are consistent across different embedding spaces (columns) and different datasets (rows). In the scatterplots, each point represents a source-target domain pair. For each plot, we also show the best-fit line and report Spearman's rank correlation coefficient $(\rho)$ and the coefficient of determination $(\mathcal{R}^2)$.}
\label{fig:trend_plots_all}
\end{figure*}

\begin{table*}[t!]
\centering
\small
\setlength{\tabcolsep}{6pt}
\begin{tabular}{lcccccccc}
\toprule
\multirow{2}{*}{\textbf{Embedding modalities}} & 
\multicolumn{2}{c}{\textbf{Geo-YFCC-Image}} & 
\multicolumn{2}{c}{\textbf{Geo-YFCC-Text}} & 
\multicolumn{2}{c}{\textbf{FMoW-Wilds}} &
\multicolumn{2}{c}{\textbf{GeoDE}}\\
\cmidrule(lr){2-3} \cmidrule(lr){4-5} \cmidrule(lr){6-7} \cmidrule(lr){8-9}
 & $|\rho|$ & $R^2$ & $|\rho|$ & $R^2$ & $|\rho|$ & $R^2$ & $|\rho|$ & $R^2$ \\
\midrule
\rowcolor{gray!15}
ResNet50/BERT & 0.71 & 0.51 & 0.40 & 0.15 & 0.46 & 0.14  & 0.34 & 0.19  \\
\addlinespace[0.3em]
GeoCLIP & 0.70 & 0.49 & 0.54 & 0.31 & 0.65 & 0.41 & 0.33 & 0.08  \\
\rowcolor{gray!15}
    \hspace{1em}ResNet50/BERT + GeoCLIP & 0.74 & 0.56 & 0.58 & 0.34 & 0.62 & 0.29 & 0.38  & 0.15  \\
\addlinespace[0.3em]
SatCLIP & 0.55 & 0.30 & 0.48 & 0.22 & 0.49 & 0.25 & 0.24 & 0.07 \\
\rowcolor{gray!15}
\hspace{1em}ResNet50/BERT + SatCLIP & 0.67 & 0.45 & 0.50 & 0.24 & 0.56  & 0.28  & 0.28 & 0.10  \\
\addlinespace[0.3em]
Geodesic & 0.42 & 0.18 & 0.42 & 0.16 & 0.24 & 0.09 & 0.17 & 0.02 \\
\rowcolor{gray!15}
\hspace{1em}ResNet50/BERT + Geodesic & 0.59 & 0.31 & 0.45 & 0.17 & 0.36 & 0.15 & 0.25 & 0.05 \\
\bottomrule
\end{tabular}
\caption{\textbf{Correlation of single-modality and multi-modality \textsc{GeoSpOT} distances.} Spearman correlation magnitude ($|\rho|$) and coefficient of determination ($R^2$) measure how well variation in \textsc{GeoSpOT} distances explains variation in zero-shot source-target domain transfer difficulty ($\Delta_{\mathcal{D}_s, \mathcal{D}_t}$ defined in \cref{equation:test_acc_relative_change}). 
Rows correspond to different combinations of data modalities used to compute the \textsc{GeoSpOT} distances. 
Indented rows are those for which two distinct data modalities are used, i.e., $ 0< \lambda < 1$ in \cref{eqn:ground dist all}. 
}

\label{tab:embedding_comparison}
\end{table*}


\Cref{fig:trend_plots_all} compares how well \textsc{GeoSpOT} distances computed from single modalities explain transfer degradation across source-target domain pairs. All plots exhibit negative correlation, i.e., larger \textsc{GeoSpOT} distances are associated with larger drops in test accuracy of target domains, consistently across datasets. For the Geo-YFCC-Image and GeoDE datasets, \textsc{GeoSpOT} with ResNet50 embeddings yields the highest magnitude $\rho$ across the embedding types ($|\rho| \approx 0.71$ and $|\rho| \approx 0.34$, respectively). For these two datasets, this is closely matched by \textsc{GeoSpOT} with GeoCLIP embeddings, with $|\rho|\approx 0.70 $ for Geo-YFCC-Image and $|\rho|\approx 0.33$ for GeoDE. Notably, \textsc{GeoSpOT} with GeoCLIP embeddings achieves the highest rank correlation in Geo-YFCC-Text and FMoW-Wilds ($|\rho| \approx 0.54$ and $|\rho| \approx 0.65$, respectively), indicating that GeoCLIP-based representations generalize well across modalities.

\textsc{GeoSpOT} with GeoCLIP embeddings consistently serves as a reliable predictor of $\Delta_{\mathcal{D}_s, \mathcal{D}_t}$ across datasets, despite depending solely on location information. Notably, as a model pretrained on geotagged Flickr images, GeoCLIP still tracks transfer difficulty well in FMoW-Wilds, even though the task is predicting land use category from satellite images. This may suggest that GeoCLIP models encode human and environmental patterns that correlate with domain shifts in remote-sensing imagery through training on globally distributed and semantically meaningful data.

In comparison, \textsc{GeoSpOT} distances based on SatCLIP embeddings are informative but weaker (e.g., $|\rho|\approx0.55$ for Geo-YFCC-Image), because SatCLIP is pretrained with satellite imagery, a much more distinct image modality to Geo-YFCC samples than Flickr images, which GeoCLIP is pretrained on. By contrast, \textsc{GeoSpOT} results on geodesic distances demonstrate the weakest predictability (e.g., $|\rho|\approx 0.42$ for both Geo-YFCC-Image and for Geo-YFCC-Text), and with the geodesic metric, many pairs cluster at high \textsc{GeoSpOT} distances yet span a wide spread over performance change. This suggests purely geographic distances are much less suitable to predict performance drops for domain transfer, whereas \textsc{GeoSpOT} distances derived from semantically meaningful embedding spaces provide a stronger proxy for domain shifts.

 
\Cref{tab:embedding_comparison} additionally reports performance 
when we combine image/text and location modalities to compute \textsc{GeoSpOT} distances, using $\lambda=0.5$ in \cref{eqn:ground dist all} (results for additional values of $\lambda$ included in the appendix). 
Across all combinations of modalities for image datasets, the inclusion of ResNet50 (image) features consistently yields a stronger $|\rho|$ value than the location modality alone, indicating that image features provide useful information for domain transfer. In particular, we see that for the Geo-YFCC-Image and GeoDE datasets, using ResNet50 + GeoCLIP yields the strongest rank correlation for this dataset ($|\rho| \approx 0.74$ for Geo-YFCC-Image, $|\rho| \approx 0.38$ for GeoDE), which is notably stronger than either modality alone. 

In contrast, when  location embeddings are informative for the task at hand, such as embeddings generated by GeoCLIP, the resulting location information complements the image information, as we conclude from a higher $|\rho|$ value from ResNet50 + GeoCLIP. However, when location embeddings are less representative of spatial similarity, or less relevant for the task, incorporating them does not boost predictive performance.
\begin{figure*}[tbh]
\centering
  \includegraphics[width=.95\linewidth]{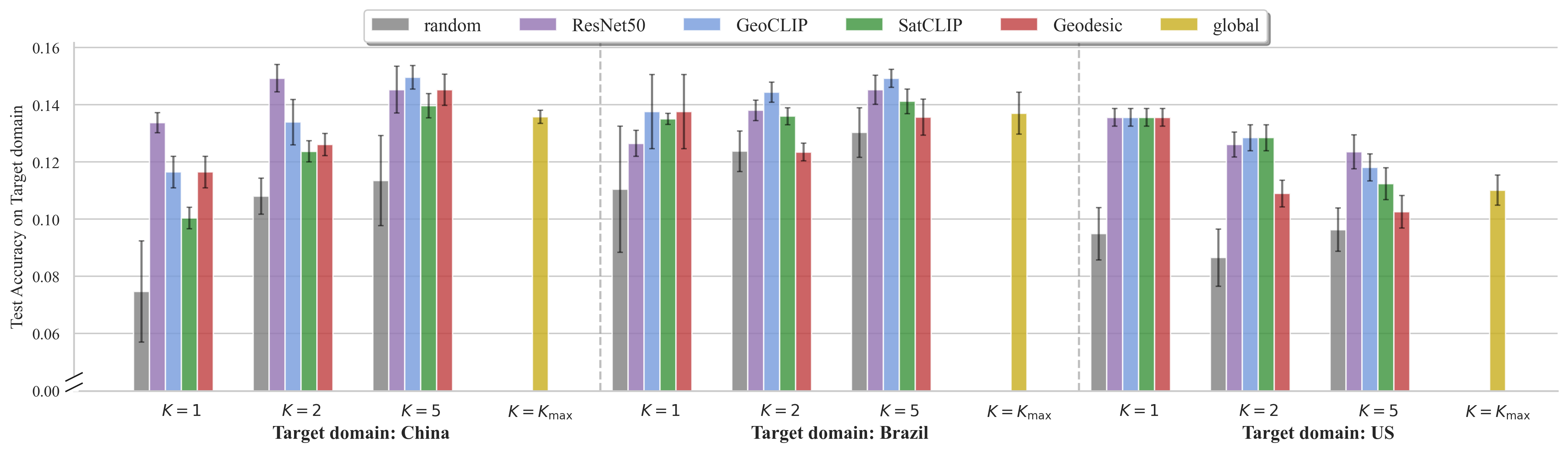}

\caption{\textbf{\textsc{GeoSpOT} distances can guide design of training datasets that transfer well to a given target domain.} We show zero-shot performance in the target domain, with models pretrained on subsets of size $N=2,000$, sampling train locations most similar to the target domain based on different \textsc{GeoSpOT} distances. Results are shown for the Geo-YFCC-Image dataset.}
\label{fig:data_selection}
\end{figure*}

\section{Constrained Dataset Selection}
\label{sec: dataset selection}

We investigate whether \textsc{GeoSpOT} distances can guide data subset selection to improve model performance under a constrained training budget, a setting that  mirrors common challenges in real-world geospatial ML deployment. Our central question is whether \textsc{GeoSpOT} distances can inform the choice of source domains for a given target. Intuitively, source domains with smaller \textsc{GeoSpOT} distances to the target domain are semantically more related, and thus training on these domains should yield models that generalize more effectively to the target distribution. 


 \setlength{\algomargin}{.5em}
\begin{algorithm}[t]
\caption{Source domain selection 
}
\label{alg:greedy_ot_selection}
\DontPrintSemicolon

\KwIn{Source 
      $\{\mathcal{D}^s_1,\dots,\mathcal{D}^s_N\}$ and target $\mathcal{D}^t$ datasets;
      number of source domains to select $K \le N$.}
\KwOut{Set of source domains $\mathcal{S}_K$.}

$\mathcal{I}_0 \gets \lceil N\rceil$; \ \
$\mathcal{S}_0 \gets \emptyset$\;
\For{$t = 1$ \KwTo $K$}{
    $i_t \gets \argmin\limits_{i \in \mathcal{I}_{t-1}}
    \textsc{GeoSpOT }\!\big(\mathcal{S}_{t-1} \cup \{\mathcal{D}^s_i\},\, \mathcal{D}^t\big)$\;
    $\mathcal{I}_t \gets \mathcal{I}_{t-1} \setminus \{i_t\}$; \ \
    $\mathcal{S}_t \gets \mathcal{S}_{t-1} \cup \{\mathcal{D}^s_{i_t}\}$\;
}
\end{algorithm}

To test this hypothesis, we design a greedy subset selection algorithm (\cref{alg:greedy_ot_selection}) that selects $K$ source domains by iteratively  minimizing the combined \textsc{GeoSpOT} distances to the target domain. At iteration $t$, the algorithm adds the domain that, when pooled with the $t-1$ previously selected domains, yields the smallest overall \textsc{GeoSpOT} distance to the target. After selecting $K$ domains, we uniformly sample equal amounts of training data from each source domain until the budget (e.g., $2,000$ samples) is reached. The model is then trained on the combined set, as described in \Cref{sec:embeddings_model_training} and evaluated on the target domain's test set. We compare against two baselines under the same training budget: (1) \textit{global training}, where the model uses data from all non-target domains (equivalent to setting $K$ to the total number of source domains), and (2) \textit{random subset training}, where $K$ source domains are selected uniformly at random and sampled in equal proportions.

We evaluate \cref{alg:greedy_ot_selection} on Geo-YFCC-Image and Geo-YFCC-Text datasets under varying conditions of data scarcity. \Cref{fig:data_selection} shows the results for three target countries (China, Brazil, and the United States) using $K \in \{1, 2, 5\}$ source domains under a total budget of $N=2,000$ data points for Geo-YFCC-Image. Results for Geo-YFCC-Text and for other budgets are provided in the Appendix.
All experiments are run for five different random seeds for data sampling; we report the mean test accuracy across runs. For the random-subset baseline, the seeds are additionally used to uniformly sample $K$ source domains. \looseness-1

\begin{figure}[t] 
\centering
\begin{subfigure}{\columnwidth}
  \centering
  \includegraphics[trim={0em 1em 0em 0em}, clip, width=.95\linewidth]{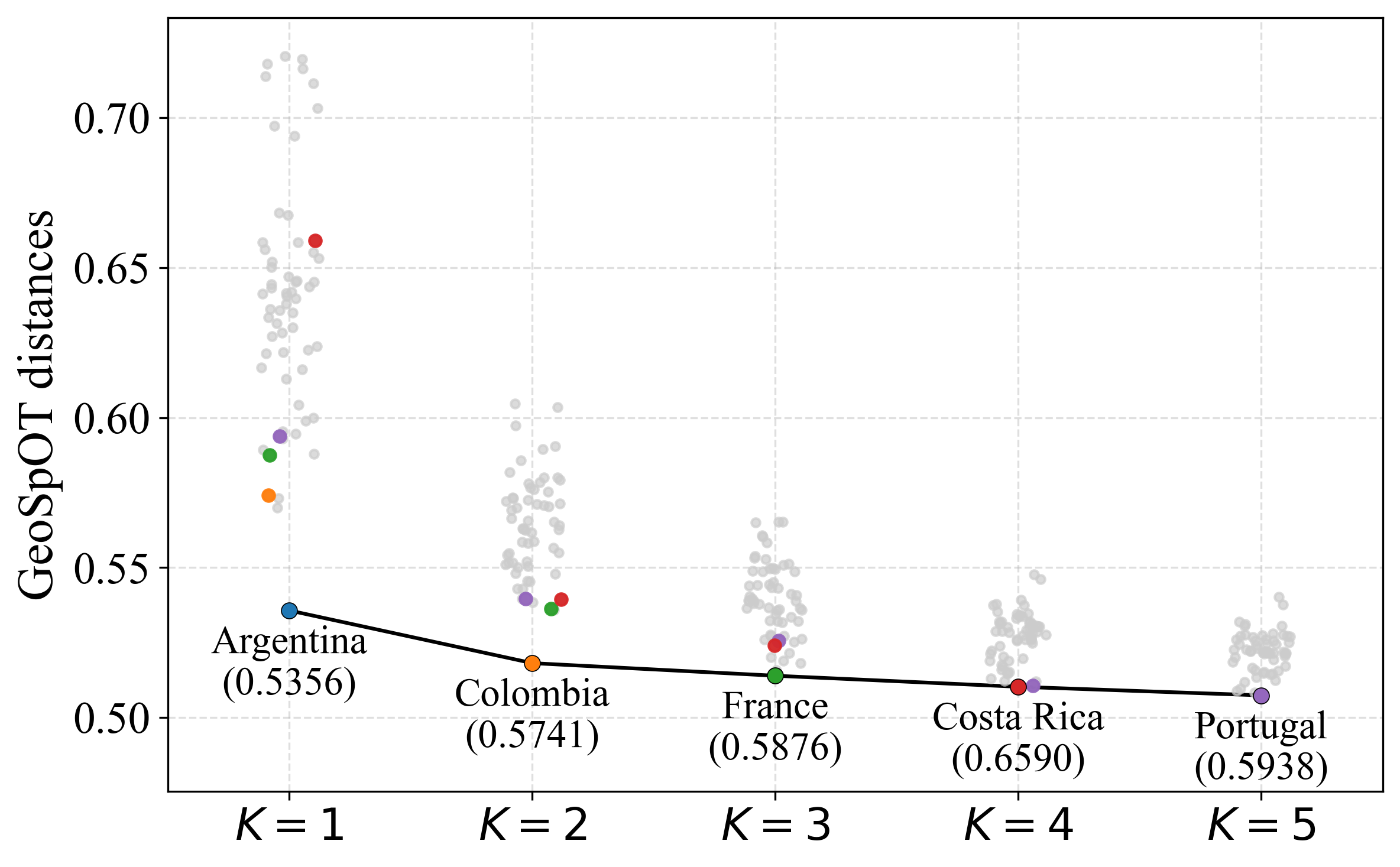}
  \label{subfig:geoclip_selected_country_brazil}
\end{subfigure}

\caption{\textbf{Complementary---not nearest---domains form optimal pools that jointly minimize the combined \textsc{GeoSpOT} distance.} Source domains selected by \cref{alg:greedy_ot_selection} using GeoCLIP-based \textsc{GeoSpOT} distances for target domain Brazil in Geo-YFCC-Image dataset. Points represent the combined \textsc{GeoSpOT} distances if that country were to be added to the source domains chosen at previous $K$ values. Values below each selected domain show their individual \textsc{GeoSpOT} distances to Brazil when $K=1$. 
}
\label{fig:selected_country}
\end{figure}

Across all target domains, selecting source domains according to any \textsc{GeoSpOT} distance consistently outperforms random selection. For China, domains chosen based on the ResNet50-based distances yield the best performance for $K=1, 2$, and the strongest \textsc{GeoSpOT}-based method surpasses the global subset at $K=2, 5$. Increasing $K$ improves performance for China but offers only marginal gains for Brazil, and, interestingly, hurts performance for the United States. For Brazil, \textsc{GeoSpOT}-GeoCLIP  performs best overall, matching the geodesic-based selection at $K=1$. For the United States, the closest domain under all distances is Canada, hence the matching results; for $K=2$, \textsc{GeoSpOT}-GeoCLIP selection achieves the highest accuracy, while for $K=5$, \textsc{GeoSpOT}-ResNet50 performs best, with both outperforming the global subset.

 \Cref{fig:selected_country} visualizes the greedily selected domains obtained by minimizing the combined \textsc{GeoSpOT} distance to Brazil. GeoCLIP-based \textsc{GeoSpOT} distance identifies not only geographically close domains (e.g., Argentina, Colombia), but also some that are geographically distant but likely related in other ways (e.g., France, Portugal). Notably, the domains that minimize the combined distance are not necessarily the individually closest to the target; rather, they are collectively complementary in distribution, such that their union minimizes the overall \textsc{GeoSpOT} distance.

\section{Applicability Maps}
\label{sec:visualizing aoas}

\begin{figure}[t]
\centering
\includegraphics[width=\linewidth]{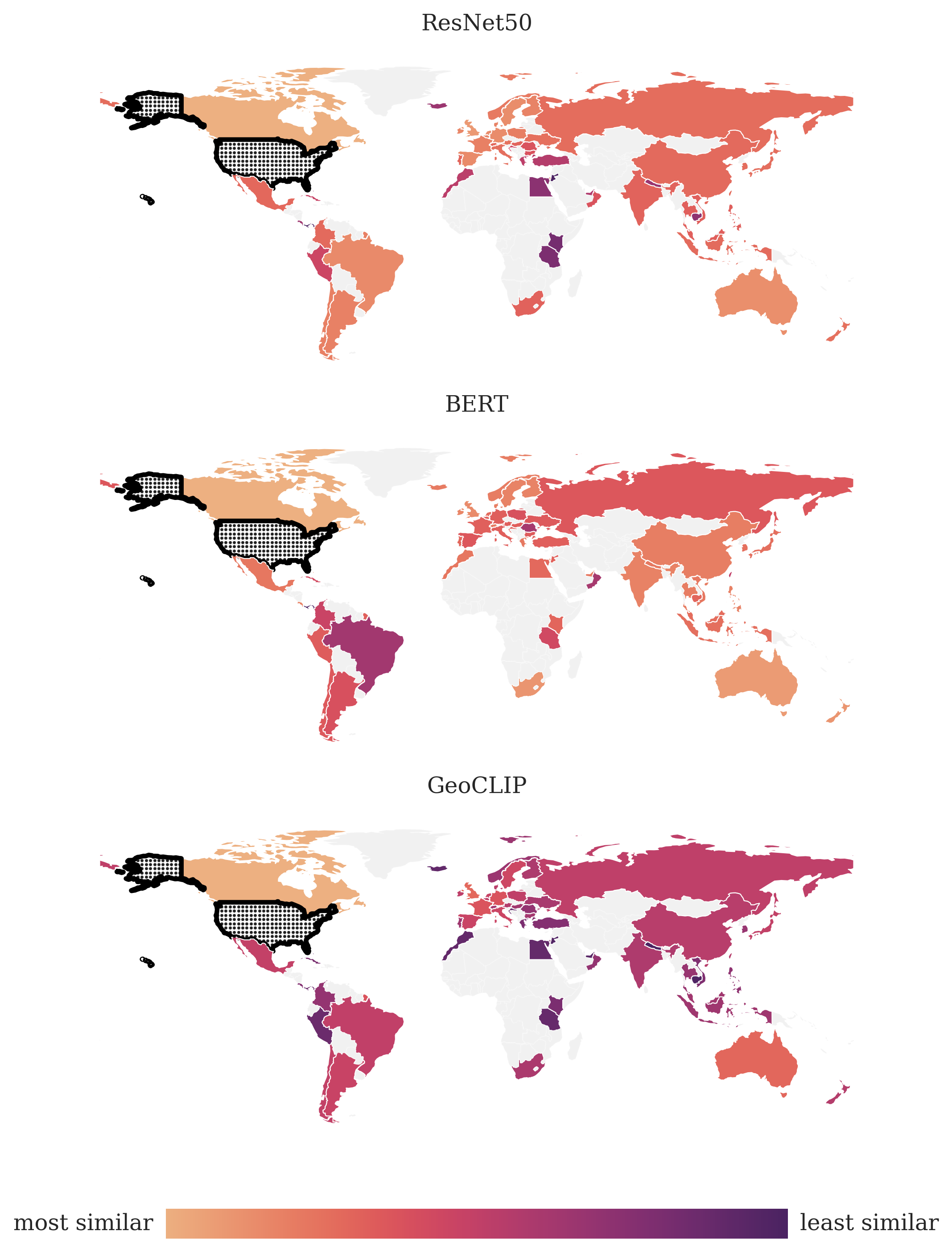}
\caption{\textbf{Visualizing \textsc{GeoSpOT} distances can identify relevant domains for data-sourcing or deployment.} Applicability maps for the United States (black, dotted) for different types of \textsc{GeoSpOT} distances on the Geo-YFCC Dataset. Color scales are normalized within each panel.
}
\label{fig:aoa_maps}
\end{figure}

\textsc{GeoSpOT} distances exhibit strong explanatory power for cross-domain performance. To better understand their structure, we visualize them as 
global \textit{applicability maps}: geographic heatmaps colored by the relative \textsc{GeoSpOT} distance from a reference domain (\Cref{fig:aoa_maps}). Regions with lower values (yellow/orange) correspond to domains that are closer in the joint feature-location space and thus more similar or transferable. 

These maps support two complementary interpretations: 
\begin{enumerate}
    \item \textit{Transfer applicability}: Given a model trained on a given domain, identify other spatial domains in which the model is expected to maintain its predictive performance
    \item \textit{Data sourcing}: Given a target domain of interest, 
    identify other source domains whose data are likely to provide the most effective training signal for the target.
\end{enumerate}

With \textsc{GeoSpOT} distances, a practitioner who, for example, has a model trained in the United States could use \Cref{fig:aoa_maps} to predict that their model could reliably transfer to countries such as Canada and Australia. A practitioner who wants to tackle a task within the United States, but has no training data available there, might also look at \Cref{fig:aoa_maps} to determine which other countries to obtain training data from, to train a model that transfers best to the US. These  applicability maps provide an interpretable, data-driven visualization of model generalization potential across geography. Derived from \textsc{GeoSpOT} distances and supported by the experimental evidence in \Cref{sec: predict w dist,sec: dataset selection}, they offer a principled way to assess where a model trained in one region can be reliably deployed, and which regions are most promising sources of transferable data for new tasks.


\section{Discussion}
\label{section:conclusion} 

In this work, we introduced GeoSpatial Optimal Transport (\textsc{GeoSpOT}), a geographically aware distance measure that combines feature and location information to quantify dissimilarity between geospatial domains.
Across four diverse geospatial datasets, \textsc{GeoSpOT} distances consistently predict cross-domain transfer difficulty.  Distances based on image/text and pretrained location embeddings show substantially stronger explanatory power than raw geodesic distance alone. The performance of \textsc{GeoSpOT} distances based on location embeddings alone is particularly noteworthy, demonstrating that meaningful, task-agnostic estimates of geospatial domain transferability can be obtained  \emph{even before any observations, features, or labels} are collected. 
Beyond prediction, we show that \textsc{GeoSpOT} can guide training data selection for zero-shot domain adaptation and visualize where models are likely to transfer successfully---or benefit most from additional data---before any data acquisition occurs.

Our findings open several directions for future work. We focus on image and text classification, but extending \textsc{GeoSpOT} to a broader range of  prediction tasks could further test its generality. In addition, our dataset selection experiments primarily demonstrate feasibility; future work could refine the selection algorithm to improve performance and scalability. 

\section*{Acknowledgements}

This material is based upon work supported by the NSF
Graduate Research Fellowship under Grant No. DGE 2040434. We would like to acknowledge use of Jetstream2 at Indiana University through allocation CIS240692 from the Advanced Cyberinfrastructure Coordination Ecosystem: Services \& Support (ACCESS) program, which is supported by National Science Foundation grants \#2138259, \#2138286, \#2138307, \#2137603, and \#2138296. DAM acknowledges the Chan Zuckerberg Initiative Foundation for establishing the
Kempner Institute for the Study of Natural and Artificial Intelligence, and NSF Award No. 2229881, NSF AI Institute for Societal Decision Making (NSF AI-SDM).

\bibliography{main}

@String(CVPR= {IEEE Conf. Comput. Vis. Pattern Recog.})

@String(ICCV= {Int. Conf. Comput. Vis.})

@String(ICIP = {IEEE Int. Conf. Image Process.})

@String(ICLR = {Int. Conf. Learn. Represent.})

@String(AAAI = {AAAI})

@String(CVPR  = {CVPR})

@String(ICCV  = {ICCV})

@String(ICIP  = {ICIP})

@String(ICLR  = {ICLR})

@inproceedings{wilds2021,
  title = {{WILDS}: A Benchmark of in-the-Wild Distribution Shifts},
  author = {Pang Wei Koh and Shiori Sagawa and Henrik Marklund and Sang Michael Xie and Marvin Zhang and Akshay Balsubramani and Weihua Hu and Michihiro Yasunaga and Richard Lanas Phillips and Irena Gao and Tony Lee and Etienne David and Ian Stavness and Wei Guo and Berton A. Earnshaw and Imran S. Haque and Sara Beery and Jure Leskovec and Anshul Kundaje and Emma Pierson and Sergey Levine and Chelsea Finn and Percy Liang},
  booktitle = {International Conference on Machine Learning (ICML)},
  year = {2021}
}

@inproceedings{fmow2018,
  title={Functional Map of the World},
  author={Christie, Gordon and Fendley, Neil and Wilson, James and Mukherjee, Ryan},
  booktitle={CVPR},
  year={2018}
}

@article{klemmer2023satclip,
  title={{SatCLIP}: Global, General-Purpose Location Embeddings with Satellite Imagery},
  author={Klemmer, Konstantin and Rolf, Esther and Robinson, Caleb and Mackey, Lester and Ru{\ss}wurm, Marc},
  journal={AAAI},
  year={2025}
}

@inproceedings{rolf2024missioncriticalsatellite,
author = {Rolf, Esther and Klemmer, Konstantin and Robinson, Caleb and Kerner, Hannah},
title = {Position: mission critical - satellite data is a distinct modality in machine learning},
year = {2024},
publisher = {JMLR.org},
booktitle = {Proceedings of the 41st International Conference on Machine Learning},
articleno = {1737},
numpages = {16},
location = {Vienna, Austria},
series = {ICML'24}
}

@inproceedings{federici2021distributionshift,
 author = {Federici, Marco and Tomioka, Ryota and Forr\'{e}, Patrick},
 booktitle = {Advances in Neural Information Processing Systems},
 editor = {M. Ranzato and A. Beygelzimer and Y. Dauphin and P.S. Liang and J. Wortman Vaughan},
 pages = {17628--17641},
 publisher = {Curran Associates, Inc.},
 title = {An Information-theoretic Approach to Distribution Shifts},
 url = {https://proceedings.neurips.cc/paper_files/paper/2021/file/93661c10ed346f9692f4d512319799b3-Paper.pdf},
 volume = {34},
 year = {2021}
}

@INPROCEEDINGS{he2015resnet,
  author={He, Kaiming and Zhang, Xiangyu and Ren, Shaoqing and Sun, Jian},
  booktitle={2016 IEEE Conference on Computer Vision and Pattern Recognition (CVPR)}, 
  title={Deep Residual Learning for Image Recognition}, 
  year={2016},
  volume={},
  number={},
  pages={770-778},
  doi={10.1109/CVPR.2016.90}}

@article{dosovitskiy2020vit,
  title={An Image is Worth 16x16 Words: Transformers for Image Recognition at Scale},
  author={Dosovitskiy, Alexey and Beyer, Lucas and Kolesnikov, Alexander and Weissenborn, Dirk and Zhai, Xiaohua and Unterthiner, Thomas and  Dehghani, Mostafa and Minderer, Matthias and Heigold, Georg and Gelly, Sylvain and Uszkoreit, Jakob and Houlsby, Neil},
  journal={ICLR},
  year={2021}
}

@article{Ansarifar2021,
  author = {Ansarifar, Javad and Wang, Lizhi and Archontoulis, Sotirios V.},
  title = {An interaction regression model for crop yield prediction},
  journal = {Scientific Reports},
  volume = {11},
  number = {1},
  pages = {17754},
  year = {2021},
  doi = {10.1038/s41598-021-97221-7},
  url = {https://doi.org/10.1038/s41598-021-97221-7},
  issn = {2045-2322}
}

@Article{Linardos2022,
AUTHOR = {Linardos, Vasileios and Drakaki, Maria and Tzionas, Panagiotis and Karnavas, Yannis L.},
TITLE = {Machine Learning in Disaster Management: Recent Developments in Methods and Applications},
JOURNAL = {Machine Learning and Knowledge Extraction},
VOLUME = {4},
YEAR = {2022},
NUMBER = {2},
PAGES = {446--473},
URL = {https://www.mdpi.com/2504-4990/4/2/20},
ISSN = {2504-4990},
DOI = {10.3390/make4020020}
}

@Article{Hu2017,
  author={Hu, Ke and Rahman, Ashfaqur and Bhrugubanda, Hari and Sivaraman, Vijay},
  journal={IEEE Sensors Journal}, 
  title={{HazeEst}: Machine Learning Based Metropolitan Air Pollution Estimation From Fixed and Mobile Sensors}, 
  year={2017},
  volume={17},
  number={11},
  pages={3517-3525},
  doi={10.1109/JSEN.2017.2690975}}

@article{Sudmanns2020,
author = {Martin Sudmanns, Dirk Tiede, Hannah Augustin and Stefan Lang},
title = {Assessing global {S}entinel-2 coverage dynamics and data availability for operational {E}arth observation {(EO)} applications using the {EO-Compass}},
journal = {International Journal of Digital Earth},
volume = {13},
number = {7},
pages = {768--784},
year = {2020},
publisher = {Taylor \& Francis},
doi = {10.1080/17538947.2019.1572799},
URL = {https://doi.org/10.1080/17538947.2019.1572799},
eprint = {https://doi.org/10.1080/17538947.2019.1572799}
}

@article{alvarez2020geometric,
  title={Geometric dataset distances via optimal transport},
  author={Alvarez-Melis, David and Fusi, Nicolo},
  journal={Advances in Neural Information Processing Systems},
  volume={33},
  pages={21428--21439},
  year={2020}
}

@inproceedings{ekim2025distribution,
  title={Distribution shifts at scale: Out-of-distribution detection in earth observation},
  author={Ekim, Burak and Tadesse, Girmaw Abebe and Robinson, Caleb and Hacheme, Gilles and Schmitt, Michael and Dodhia, Rahul and Ferres, Juan M Lavista},
  booktitle={Proceedings of the Computer Vision and Pattern Recognition Conference},
  pages={2265--2274},
  year={2025}
}

@article{lynch2021leveraging,
  title={Leveraging domain adaptation for low-resource geospatial machine learning},
  author={Lynch, Jack and Wookey, Sam},
  journal={arXiv preprint arXiv:2107.04983},
  year={2021}
}

@inproceedings{nachmany2019detecting,
  title={Detecting roads from satellite imagery in the developing world},
  author={Nachmany, Yoni and Alemohammad, Hamed},
  booktitle={Proceedings of the IEEE/CVF Conference on Computer Vision and Pattern Recognition Workshops},
  pages={83--89},
  year={2019}
}

@article{shankar2017no,
  title={No classification without representation: Assessing geodiversity issues in open data sets for the developing world},
  author={Shankar, Shreya and Halpern, Yoni and Breck, Eric and Atwood, James and Wilson, Jimbo and Sculley, D},
  journal={arXiv preprint arXiv:1711.08536},
  year={2017}
}

@inproceedings{gawlikowski2021out,
  title={Out-of-Distribution Detection in Satellite Image Classification},
  author={Gawlikowski, Jakob and Saha, Sudipan and Kruspe, Anna and Zhu, Xiao Xiang},
  booktitle={RobustML Workshop at ICLR 2021},
  pages={1--5},
  year={2021},
  organization={ICLR}
}

@ARTICLE{makkar2022adv,
  author={Makkar, Nikhil and Yang, Lexie and Prasad, Saurabh},
  journal={IEEE Journal of Selected Topics in Applied Earth Observations and Remote Sensing}, 
  title={Adversarial Learning Based Discriminative Domain Adaptation for Geospatial Image Analysis}, 
  year={2022},
  volume={15},
  number={},
  pages={150-162},
  doi={10.1109/JSTARS.2021.3132259}}

@inproceedings{lin2019spatially,
  title={Spatially-aware domain adaptation for semantic segmentation of urban scenes},
  author={Lin, Yong-Xiang and Tan, Daniel Stanley and Cheng, Wen-Huang and Chen, Yung-Yao and Hua, Kai-Lung},
  booktitle={2019 IEEE International Conference on Image Processing (ICIP)},
  pages={1870--1874},
  year={2019},
  organization={IEEE}
}

@article{crasto2025robustness,
  title={Robustness to Geographic Distribution Shift using Location Encoders},
  author={Crasto, Ruth},
  journal={arXiv preprint arXiv:2503.02036},
  year={2025}
}

@inproceedings{kifer2004detecting,
  title={Detecting change in data streams},
  author={Kifer, Daniel and Ben-David, Shai and Gehrke, Johannes},
  booktitle={VLDB},
  volume={4},
  pages={180--191},
  year={2004},
  organization={Toronto, Canada}
}

@article{ben2006analysis,
  title={Analysis of representations for domain adaptation},
  author={Ben-David, Shai and Blitzer, John and Crammer, Koby and Pereira, Fernando},
  journal={Advances in neural information processing systems},
  volume={19},
  year={2006}
}

@inproceedings{achille2019task2vec,
  title={Task2{V}ec: Task embedding for meta-learning},
  author={Achille, Alessandro and Lam, Michael and Tewari, Rahul and Ravichandran, Avinash and Maji, Subhransu and Fowlkes, Charless C and Soatto, Stefano and Perona, Pietro},
  booktitle={Proceedings of the IEEE/CVF international conference on computer vision},
  pages={6430--6439},
  year={2019}
}

@inproceedings{tan2021otce,
  title={Otce: A transferability metric for cross-domain cross-task representations},
  author={Tan, Yang and Li, Yang and Huang, Shao-Lun},
  booktitle={Proceedings of the IEEE/CVF conference on computer vision and pattern recognition},
  pages={15779--15788},
  year={2021}
}

@article{meyer2021predicting,
  title={Predicting into unknown space? {E}stimating the area of applicability of spatial prediction models},
  author={Meyer, Hanna and Pebesma, Edzer},
  journal={Methods in Ecology and Evolution},
  volume={12},
  number={9},
  pages={1620--1633},
  year={2021},
  publisher={Wiley Online Library}
}

@article{courty2017joint,
  title={Joint distribution optimal transportation for domain adaptation},
  author={Courty, Nicolas and Flamary, R{\'e}mi and Habrard, Amaury and Rakotomamonjy, Alain},
  journal={Advances in neural information processing systems},
  volume={30},
  year={2017}
}

@article{yurochkin2019hierarchical,
  title={Hierarchical optimal transport for document representation},
  author={Yurochkin, Mikhail and Claici, Sebastian and Chien, Edward and Mirzazadeh, Farzaneh and Solomon, Justin M},
  journal={Advances in neural information processing systems},
  volume={32},
  year={2019}
}

@article{thomee2016yfcc,
author = {Thomee, Bart and Shamma, David A. and Friedland, Gerald and Elizalde, Benjamin and Ni, Karl and Poland, Douglas and Borth, Damian and Li, Li-Jia},
title = {{YFCC100M}: the new data in multimedia research},
year = {2016},
issue_date = {February 2016},
publisher = {Association for Computing Machinery},
address = {New York, NY, USA},
volume = {59},
number = {2},
issn = {0001-0782},
url = {https://doi.org/10.1145/2812802},
doi = {10.1145/2812802},
journal = {Commun. ACM},
month = jan,
pages = {64–73},
numpages = {10}
}

@InProceedings{dubey2021geoyfcc,
  title={Adaptive Methods for Real-World Domain Generalization},
  author={Dubey, Abhimanyu and Ramanathan, Vignesh and Pentland, Alex and Mahajan, Dhruv},
  booktitle = {IEEE/CVF Conference on Computer Vision and Pattern Recognition (CVPR)},
  month = {June},
  year = {2021}
}

@inproceedings{ramaswamy2022geode,
    author = {Vikram V. Ramaswamy and Sing Yu Lin and Dora Zhao and Aaron B. Adcock and Laurens van der Maaten and Deepti Ghadiyaram and 
              Olga Russakovsky},
    title = {{GeoDE}: a Geographically Diverse Evaluation Dataset for Object Recognition},
    booktitle = {NeurIPS Datasets and Benchmarks},
    year = {2023}
}

@inproceedings{geoclip,
  title={{GeoCLIP}: Clip-Inspired Alignment between Locations and Images for Effective Worldwide Geo-localization},
  author={Vivanco, Vicente and Nayak, Gaurav Kumar and Shah, Mubarak},
  booktitle={Advances in Neural Information Processing Systems},
  year={2023}
}

@inproceedings{devlin-etal-2019-bert,
    title = "{BERT}: Pre-training of Deep Bidirectional Transformers for Language Understanding",
    author = "Devlin, Jacob  and
      Chang, Ming-Wei  and
      Lee, Kenton  and
      Toutanova, Kristina",
    editor = "Burstein, Jill  and
      Doran, Christy  and
      Solorio, Thamar",
    booktitle = "Proceedings of the 2019 Conference of the North {A}merican Chapter of the Association for Computational Linguistics: Human Language Technologies, Volume 1 (Long and Short Papers)",
    month = jun,
    year = "2019",
    address = "Minneapolis, Minnesota",
    publisher = "Association for Computational Linguistics",
    url = "https://aclanthology.org/N19-1423/",
    doi = "10.18653/v1/N19-1423",
    pages = "4171--4186",
}

@article{tobler1970computer,
  title={A Computer Movie Simulating Urban Growth in the Detroit Region},
  author={Tobler, Waldo R.},
  journal={Economic Geography},
  volume={46},
  number={sup1},
  pages={234--240},
  year={1970},
  publisher={Taylor \& Francis},
  doi={10.2307/143141},
  url={https://doi.org/10.2307/143141}
}

@article{redko2020co,
  title={Co-optimal transport},
  author={Redko, Ievgen and Vayer, Titouan and Flamary, R{\'e}mi and Courty, Nicolas},
  journal={Advances in neural information processing systems},
  volume={33},
  number={17559-17570},
  pages={2},
  year={2020}
}

@article{yao2023improving,
  title={Improving domain generalization with domain relations},
  author={Yao, Huaxiu and Yang, Xinyu and Pan, Xinyi and Liu, Shengchao and Koh, Pang Wei and Finn, Chelsea},
  journal={arXiv preprint arXiv:2302.02609},
  year={2023}
}

@inproceedings{le2024detecting,
  title={Detecting out-of-distribution {E}arth observation images with diffusion models},
  author={Le Bellier, Georges and Audebert, Nicolas},
  booktitle={Proceedings of the IEEE/CVF Conference on Computer Vision and Pattern Recognition},
  pages={481--491},
  year={2024}
}

@inproceedings{dimitric2023nearest,
  title={Nearest Neighbor Based Out-of-Distribution Detection in Remote Sensing Scene Classification},
  author={Dimitri{\'c}, Dajana and Risojevi{\'c}, Vladimir and Simi{\'c}, Mitar},
  booktitle={2023 22nd International Symposium INFOTEH-JAHORINA (INFOTEH)},
  pages={1--4},
  year={2023},
  organization={IEEE}
}

@article{fang2022confident,
  title={Confident learning-based domain adaptation for hyperspectral image classification},
  author={Fang, Zhuoqun and Yang, Yuexin and Li, Zhaokui and Li, Wei and Chen, Yushi and Ma, Li and Du, Qian},
  journal={IEEE Transactions on Geoscience and Remote Sensing},
  volume={60},
  pages={1--16},
  year={2022},
  publisher={IEEE}
}

@article{ismael2023unsupervised,
  title={Unsupervised domain adaptation for the semantic segmentation of remote sensing images via one-shot image-to-image translation},
  author={Ismael, Sarmad F and Kayabol, Koray and Aptoula, Erchan},
  journal={IEEE Geoscience and Remote Sensing Letters},
  volume={20},
  pages={1--5},
  year={2023},
  publisher={IEEE}
}

@inproceedings{redko2019optimal,
  title={Optimal transport for multi-source domain adaptation under target shift},
  author={Redko, Ievgen and Courty, Nicolas and Flamary, R{\'e}mi and Tuia, Devis},
  booktitle={The 22nd International Conference on artificial intelligence and statistics},
  pages={849--858},
  year={2019},
  organization={PMLR}
}

@article{damodaran2020entropic,
  title={An entropic optimal transport loss for learning deep neural networks under label noise in remote sensing images},
  author={Damodaran, Bharath Bhushan and Flamary, R{\'e}mi and Seguy, Vivien and Courty, Nicolas},
  journal={Computer Vision and Image Understanding},
  volume={191},
  pages={102863},
  year={2020},
  publisher={Elsevier}
}

@misc{flamary2024pot,
  author = {Flamary, R{\'e}mi and Vincent-Cuaz, C{\'e}dric and Courty, Nicolas and Gramfort, Alexandre and Kachaiev, Oleksii and Quang Tran, Huy and David, Laurène and Bonet, Cl{\'e}ment and Cassereau, Nathan and Gnassounou, Th{\'e}o and Tanguy, Eloi and Delon, Julie and Collas, Antoine and Mazelet, Sonia and Chapel, Laetitia and Kerdoncuff, Tanguy and Yu, Xizheng and Feickert, Matthew and Krzakala, Paul and Liu, Tianlin and Fernandes Montesuma, Eduardo},
  title = {{POT} Python Optimal Transport (version 0.9.5)},
  url = {https://github.com/PythonOT/POT},
  year = {2024}
}

@article{flamary2021pot,
  author  = {R{\'e}mi Flamary and Nicolas Courty and Alexandre Gramfort and Mokhtar Z. Alaya and Aur{\'e}lie Boisbunon and Stanislas Chambon and Laetitia Chapel and Adrien Corenflos and Kilian Fatras and Nemo Fournier and L{\'e}o Gautheron and Nathalie T.H. Gayraud and Hicham Janati and Alain Rakotomamonjy and Ievgen Redko and Antoine Rolet and Antony Schutz and Vivien Seguy and Danica J. Sutherland and Romain Tavenard and Alexander Tong and Titouan Vayer},
  title   = {{POT}: Python Optimal Transport},
  journal = {Journal of Machine Learning Research},
  year    = {2021},
  volume  = {22},
  number  = {78},
  pages   = {1-8},
  url     = {http://jmlr.org/papers/v22/20-451.html}
}

@InProceedings{Zhu_2015_ICCV,
    title = {Aligning Books and Movies: Towards Story-Like Visual Explanations by Watching Movies and Reading Books},
    author = {Zhu, Yukun and Kiros, Ryan and Zemel, Rich and Salakhutdinov, Ruslan and Urtasun, Raquel and Torralba, Antonio and Fidler, Sanja},
    booktitle = {The IEEE International Conference on Computer Vision (ICCV)},
    month = {December},
    year = {2015}
}
\bibliographystyle{icml2026}

\clearpage
\appendix

%
%
%

\section{Optimal Transport framework}
\label{appendix:methodology}

To compute \textsc{GeoSpOT} distances, we use the Sinkhorn divergence method, an entropically regularized approximation that is computationally more efficient than solving the exact Optimal Transport problem. The entropic optimal transport problem, given regularization parameter $\epsilon$, is formulated as
\begin{align*}
    \min_{\pi \in \Pi(\alpha, \beta)} &\int_{\mathcal{Z}\times \mathcal{Z}} c_{\mathcal{Z}} (z, z') \mathrm{d}\pi(z, z') \\
    & + \epsilon \int_{\mathcal{Z}\times \mathcal{Z}} \pi (z, z') \log \pi(z, z') \mathrm{d}\pi(z, z').
\end{align*}
Sinkhorn divergence corrects for the regularization-induced bias of entropic OT by subtracting self-similarity terms, and is defined as
\begin{align*}
    \mathrm{S}_\epsilon (\alpha, \beta) = \text{OT}_\epsilon (\alpha, \beta) - \frac{1}{2} \text{OT}_\epsilon (\alpha, \alpha) - \frac{1}{2} \text{OT}_\epsilon (\beta, \beta),
\end{align*}
which satisfies $\mathrm{S}_\epsilon (\alpha, \alpha) = 0$ for all $\alpha$ and preserves several geometric properties of true OT distances.

In our experiments, we compute all OT-based distances using the POT (Python Optimal Transport) library \citep{flamary2021pot, flamary2024pot}. We set the entropic regularization parameter to $\epsilon=0.01$, and we use uniform sample weights for both the source and target domains.

For numerical stability, we apply max normalization to the cost matrix for each source-target domain pair, dividing all costs by the largest value observed among any pair of elements across the two domains.

\section{Experiment Details}

\begin{table*}[!t]
\centering
\begin{tabular}{lll}
\toprule
\textbf{Setting} & \textbf{Image Classification} & \textbf{Text Classification} \\
\midrule
\multirow{2}{*}{Learning rate} & $1\times 10^{-5}$ for backbone,  & \multirow{2}{*}{$2\times10^{-5}$} \\
& $1\times10^{-4}$ for classification head & \\
Optimizer & Adam & AdamW \\
Weight decay & $1\times 10^{-4}$ & 0.01 \\
Scheduler & Cosine annealing ($\eta_{\min}=1\times 10^{-6}$) & Cosine ($\gamma = 0.1$) \\
Epochs & 20 & 50 \\
Early stopping & Patience 5, start at epoch 10 & Patience 10, start at epoch 20 \\
\midrule
Train batch size & 64 & 64 \\
Eval batch size & 512 & 512 \\
\bottomrule
\end{tabular}
\caption{Training hyperparameters used for image and text classification tasks.}
\label{tab:hparams}
\end{table*}

\subsection{Datasets}
\label{appendix:dataset}

In this section, we discuss the preprocessing steps we apply to obtain the data used for our experiments.

\subsubsection{Geo-YFCC}

The Geo-YFCC dataset consists of more than $1.1$ million geotagged samples \citep{dubey2021geoyfcc}. Each sample includes an image, \verb|title|, \verb|description|, and \verb|user_tags|. Labels are derived from the \verb|user_tags| field, which may contain more than one label per sample. Following the methodology of \citep{dubey2021geoyfcc}, we expand this into a single-label dataset, duplicating samples that have more than one \verb|user_tag| to generate instances for each label. We use the original train and test splits and partition a portion of the training data into a validation set. Within each geographic domain, the size of train, validation, and test set is approximately 6:2:2. When splitting our data into the train-validation-test partitions, we ensure that no instance appears in more than one split to prevent data leakage.

For Geo-YFCC-Image dataset, after removing samples with unreadable images, we obtain 1,809,363 data samples in total (including duplicates from the single-label conversion).

For the Geo-YFCC-Text dataset, we compile text samples by combining the \verb|title| and \verb|description| fields. We remove any samples with no text in either of these fields. We also remove any URLs and normalize whitespace to a single whitespace. 

\subsubsection{FMoW-Wilds}

Function Map of the World is a large dataset of high-resolution satellite images collected around the globe for classification of buildings and land use \citep{fmow2018}. The Wilds benchmark further adapts this dataset into domain adaptation tasks with focus on spatial and temporal distribution shifts \citep{wilds2021}.

The FMoW-Wilds dataset contains 523,846 RGB images with associated spatiotemporal metadata, including \verb|latitude|, \verb|longitude|, \verb|country_code|, and \verb|year|. Temporally, FMoW-Wilds spans the years 2002-2018. Because we focus on geographic domains, we restrict the data to 2016-2017 to minimize the temporal distribution shifts. Spatially, FMoW-Wilds primarily defines domains at the continental level (Asia, Europe, Africa, the Americas, and Oceania). Due to substantial intra-continental variability and the limited number of continental domains, we instead define geographic domains at the country or region level, where more specialized traits and distributions may be observed beucase of geographic, cultural, and socioeconomic factors. To ensure there is sufficient training data and class coverage, we select the 25 countries or regions with the most training samples by \verb|country_code|, and this subset contains 152,196 image samples in total. 

The FMoW-Wilds dataset is originally divided into train, validation, test, and sequestered splits. We use the official FMoW-wilds train, validation, and test splits for our experiments.

\subsubsection{GeoDE}

The GeoDE dataset contains 61,940 images with location metadata, including \verb|gps_position|, \verb|region|, and \verb|ip_country| \citep{ramaswamy2022geode}. We use the country metadata to partition the dataset into 42 country-level domains. After removing countries with fewer than 1,000 samples, we retain 19 domains spanning different regions and continents around the globe, yielding a cleaned subset of 61,379 images. We then randomly split the cleaned dataset into train, validation, and test sets using a 6:2:2 ratio.

\subsection{Distance Computation}

We compute \textsc{GeoSpOT} distances using embeddings from several pretrained models to encode information from different modalities. For image embeddings, we use ResNet50 models initialized with ImageNet-1K pretrained weights \citep{he2015resnet}; for text embeddings, we use the \verb|bert-based-uncased| model, which is pretrained on BookCorpus \citep{Zhu_2015_ICCV} and English Wikipedia. For location embeddings, we use the location encoders from SatCLIP and GeoCLIP models. SatCLIP has multiple pretrained checkpoints that combine different vision backbones with varying spatial resolution $L$, where $L$ denotes the number of Legendre polynomials used in the spherical-harmonics location encoding \citep{klemmer2023satclip}. In our experiments, we use the SatCLIP checkpoint with a ResNet50 backbone and spatial resolution $L=40$, which yields higher-resolution, fine-grained location embeddings. For GeoCLIP, we use the pretrained model with ViT-L/14 vision backbone \citep{geoclip}.

\subsection{Model Training}


\paragraph{Image Classification} 
We evaluate both ResNet50 (with around 25.6 million parameters) \citep{he2015resnet} and ViT-Small (with around 22.1 million parameters) \citep{dosovitskiy2020vit} for image classification experiments, covering convolutional neural network and vision transformer architectures. For ResNet50, we initialize the model with ImageNet-1K pretrained weights from the torchvision library. For ViT-Small models, we use the \texttt{vit\_small\_patch16\_224} backbone and load model weights pretrained on ImageNet-21K and fine-tuned on ImageNet-1K.

\paragraph{Text Classification}
We train BERT models for text classification tasks. We use BertForSequenceClassification models from HuggingFace and load \texttt{bert-base-uncased} pretrained checkpoint.

We report hyperparameters used for all training experiments in \Cref{tab:hparams}.




\section{Results}

\subsection{Ablation Studies}

\begin{figure*}[!ht] 
\centering
\begin{subfigure}{\textwidth}
  \centering
  \includegraphics[width=\linewidth]{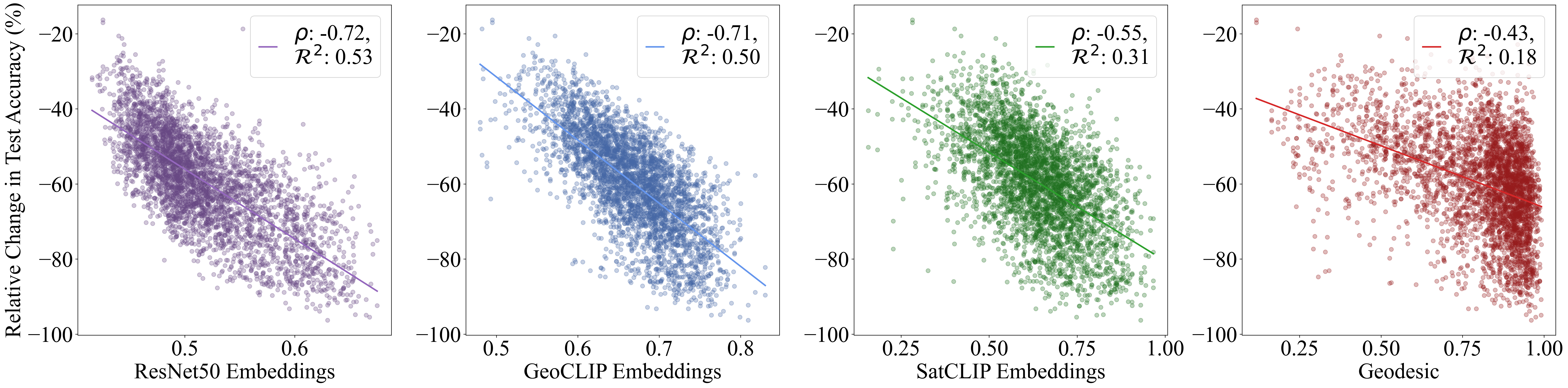}
  \caption{Geo-YFCC-Image}
  \label{subfig:trend_plots_geoyfcc_img}
\end{subfigure}

\medskip

\begin{subfigure}{\textwidth}
  \centering
  \includegraphics[width=\linewidth]{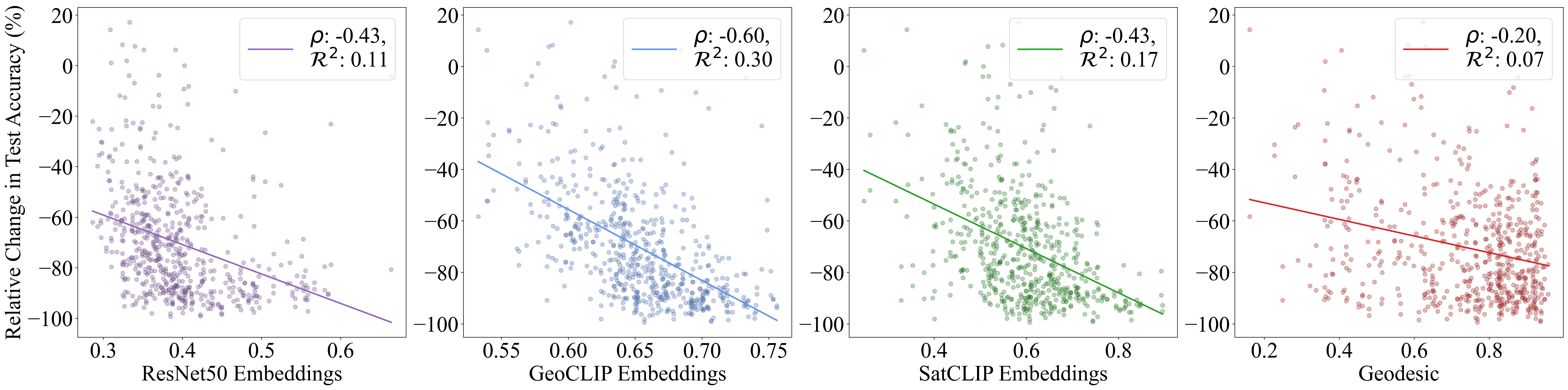}
  \caption{FMoW-Wilds}
  \label{subfig:trend_plots_fmow_wilds}
\end{subfigure}

\caption{\textbf{Correlation between \textsc{GeoSpOT} distances and zero-shot transfer performance with ViT-Small models on Geo-YFCC-Image and FMoW-Wilds.}}
\label{fig:trend_plots_vit}
\end{figure*}

We further conduct ablation studies to assess how the choice of backbone models used for domain transfer can affect our results. \Cref{fig:trend_plots_vit} reports the correlation between \textsc{GeoSpOT} distances and zero-shot transfer performance using ViT-Small models for image classification on the Geo-YFCC-Image and FMoW-Wilds datasets. The observed trends closely match with the results with ResNet50 transfer performance in \Cref{fig:trend_plots_all}.

\subsection{Combined \textsc{GeoSpOT} Distances}
\label{appx: combo ot}

\begin{figure*}[!tbh]
\centering
\includegraphics[width=\textwidth]{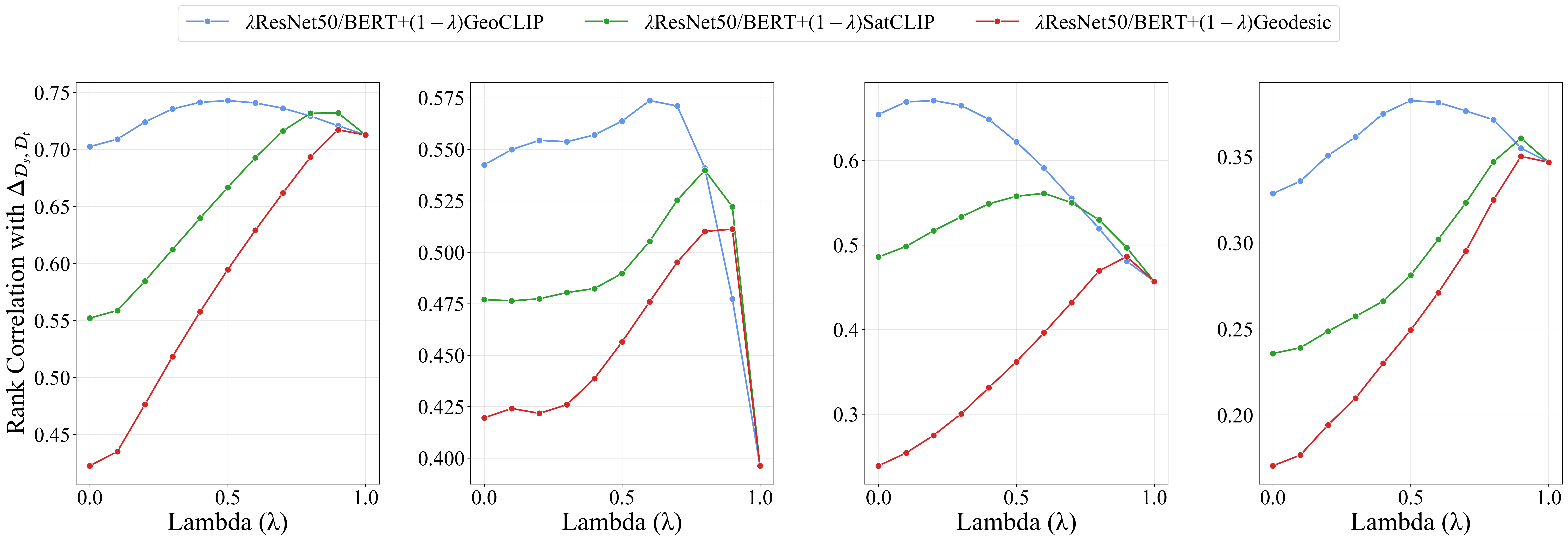}

\vspace{0.5em} 

\begin{minipage}{0.245\textwidth}
\centering
\hspace{6mm} (a) Geo-YFCC-Image
\end{minipage}%
\hfill
\begin{minipage}{0.245\textwidth}
\centering
\hspace{6mm} (b) Geo-YFCC-Text
\end{minipage}%
\hfill
\begin{minipage}{0.245\textwidth}
\centering
\hspace{6mm} (c) FMoW
\end{minipage}%
\hfill
\begin{minipage}{0.245\textwidth}
\centering
\hspace{6mm} (d) GeoDE
\end{minipage}

\caption{\textbf{Rank Correlation ($|\rho|$) across datasets as a function of the weighting parameter $\lambda$.}  
Each panel shows how $|\rho|$ changes as the image/text and location-based information are combined, weighted by $\lambda$.}
\label{fig:lambda_all_datasets}
\end{figure*}

In this section, we study the effect of varying $\lambda$ in \Cref{eqn:ground dist all}. \Cref{tab:embedding_comparison} reports results for $\lambda = 0.5$, and we also evaluate $\lambda$ at evenly spaced values between 0 and 1 in increments of 0.1, i.e., $\lambda \in \{0.0, 0.1, 0.2, \dots, 0.9, 1.0\}$.

Across all datasets, we observe that selecting an appropriate value of $\lambda$ yields an improvement in the rank correlation coefficient ($\rho$). The optimal value of $\lambda$ varies by dataset and appears to depend on the relative scale of the two distance components (the image/text embedding distances, $d_\mathcal{X}$, and the location-based distances, $d_\mathcal{L}$). For instance, in the Geo-YFCC-Text dataset, the BERT-derived embedding distances span a smaller range of values compared to the corresponding location distances (see \Cref{fig:trend_plots_all}). As a result, larger values of $\lambda$ tend to perform better. Developing principled methods for selecting an optimal $\lambda$ to improve the explanatory power of \textsc{GeoSpOT} distances is an interesting direction for future work.

\subsection{Constrained Dataset Selection}
\label{appx: dataset selection}


\begin{figure*}[!tbh]
\centering

\begin{subfigure}{0.95\linewidth}
  \centering
  \includegraphics[width=\linewidth]{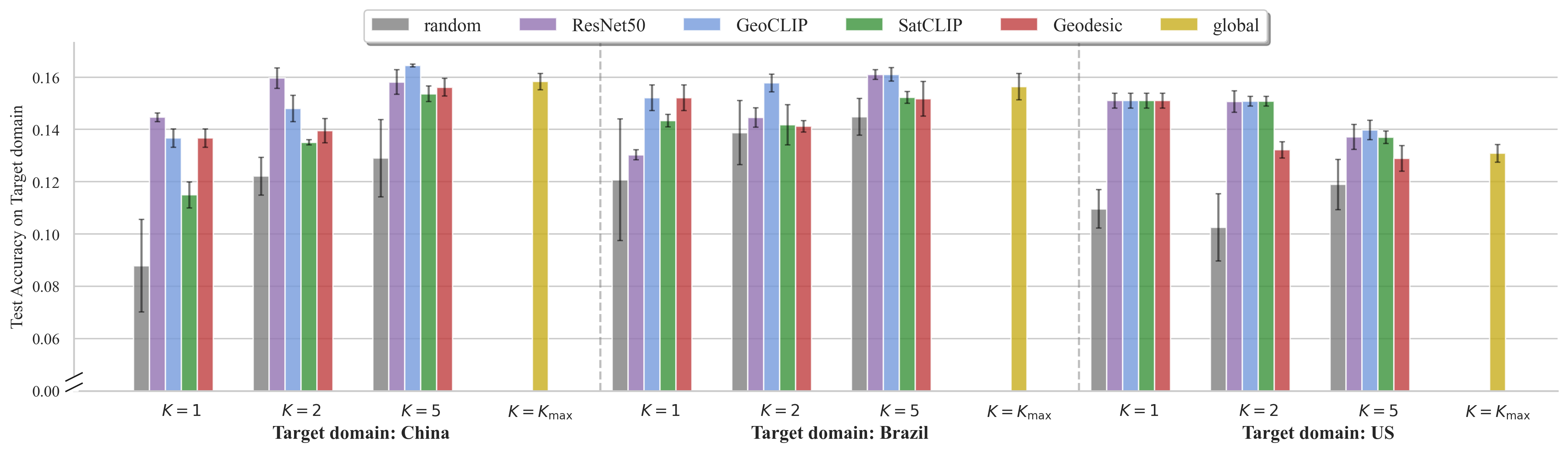}
  \caption{Budget $N=5{,}000$}
  \label{fig:data_selection_5000}
\end{subfigure}

\vspace{0.5em}

\begin{subfigure}{0.95\linewidth}
  \centering
  \includegraphics[width=\linewidth]{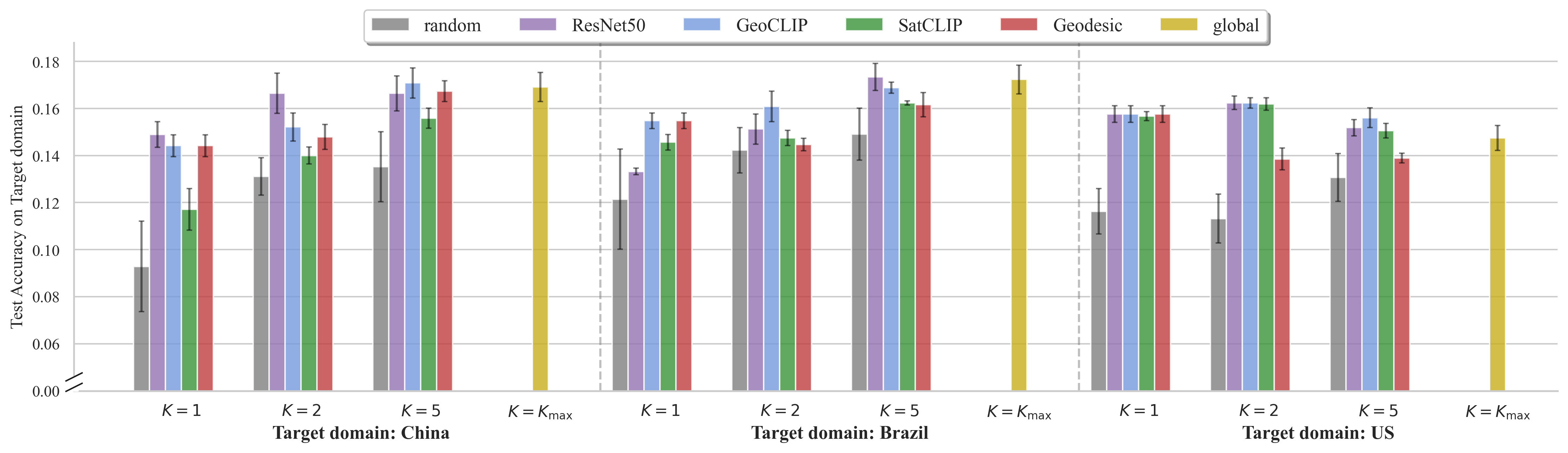}
  \caption{Budget $N=10{,}000$}
  \label{fig:data_selection_10000}
\end{subfigure}

\caption{\textbf{Constrained dataset selection for Geo-YFCC-Image:} Zero-shot performance on the target domain with models pretrained on subsets selected via different \textsc{GeoSpOT} distances.}
\label{fig:data_selection_more_budgets}
\end{figure*}

\begin{figure*}[!tbh]
\centering

\begin{subfigure}{0.95\linewidth}
  \centering
  \includegraphics[width=\linewidth]{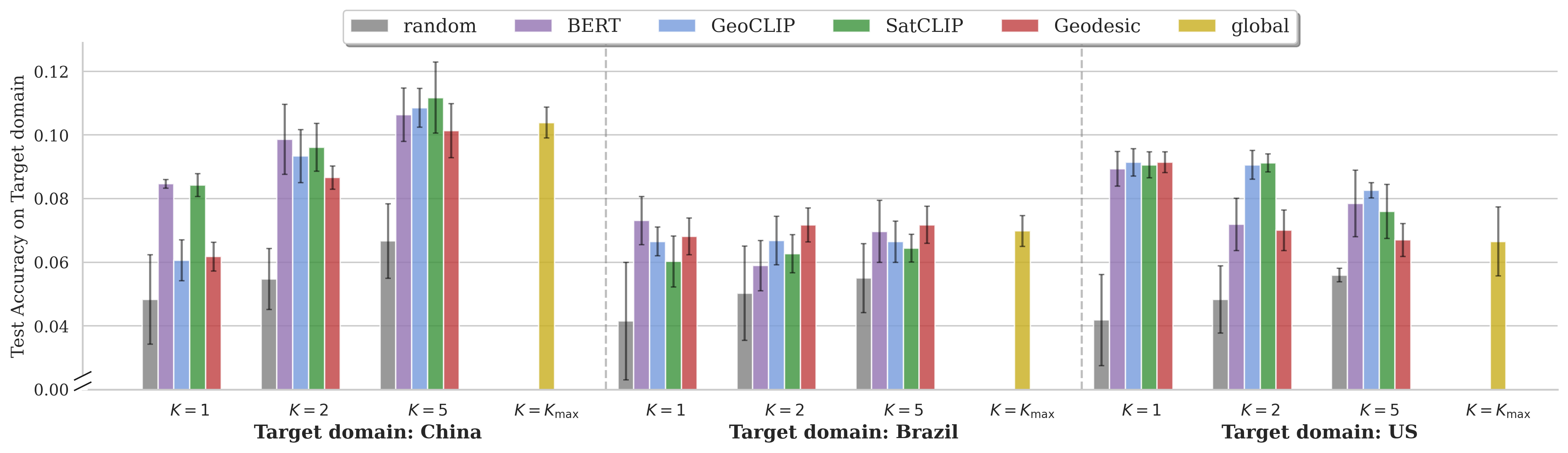}
  \caption{Budget $N=2{,}000$}
  \label{fig:text_data_selection_2000}
\end{subfigure}

\begin{subfigure}{0.95\linewidth}
  \centering
  \includegraphics[width=\linewidth]{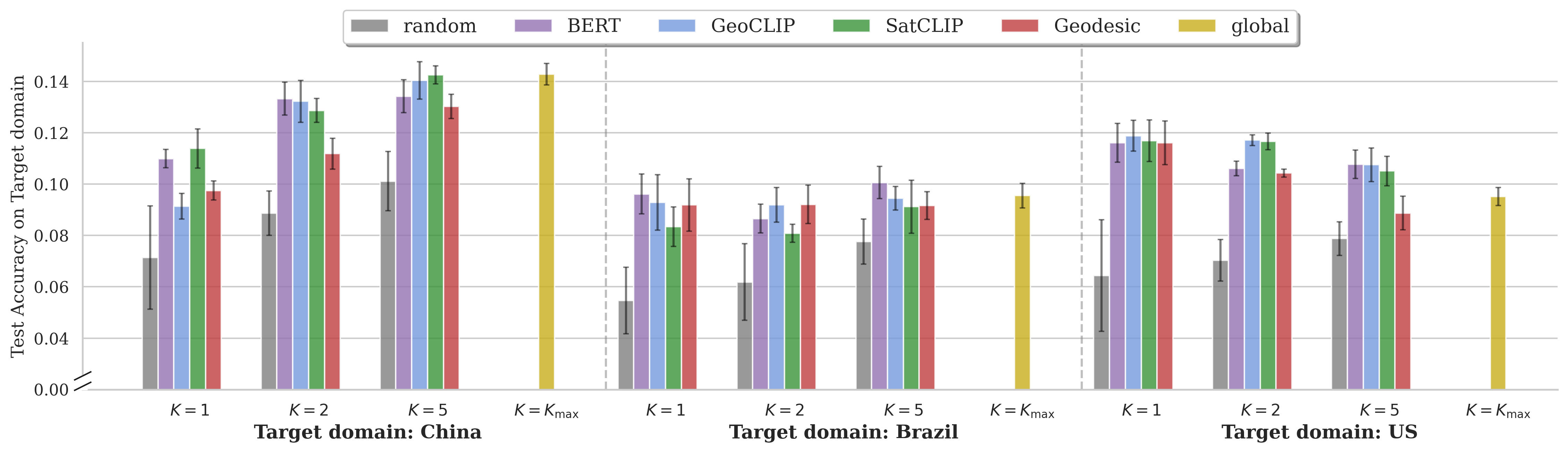}
  \caption{Budget $N=5{,}000$}
  \label{fig:text_data_selection_5000}
\end{subfigure}

\vspace{0.5em}

\begin{subfigure}{0.95\linewidth}
  \centering
  \includegraphics[width=\linewidth]{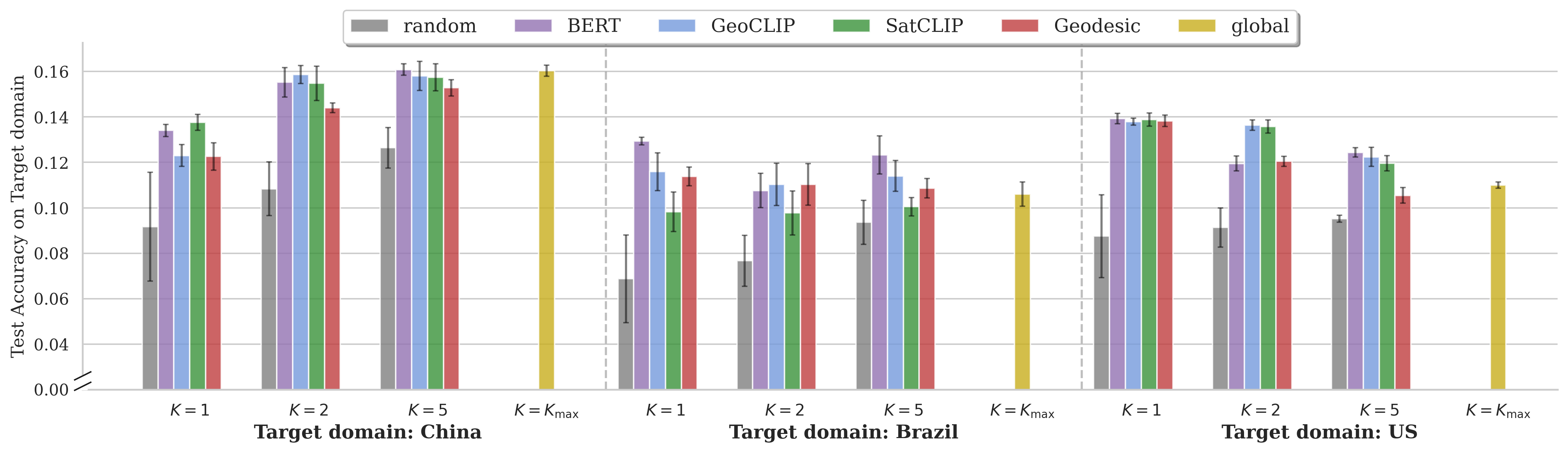}
  \caption{Budget $N=10{,}000$}
  \label{fig:text_data_selection_10000}
\end{subfigure}

\caption{\textbf{Constrained dataset selection for Geo-YFCC-Text:} Zero-shot performance on the target domain with models pretrained on subsets selected via different \textsc{GeoSpOT} distances.}
\label{fig:text_data_selection}
\end{figure*}

We provide additional results for the constrained dataset selection experiments under larger budgets. For the Geo-YFCC-Image dataset, we report results for budgets $N=5,000$ and $10,000$ in \Cref{fig:data_selection_more_budgets}, in addition to $N=2,000$ reported in \Cref{sec: dataset selection}. For the Geo-YFCC-Text dataset, we report results for budgets $N=2,000$, $5,000$ and $10,000$ in \Cref{fig:text_data_selection}.

\Cref{fig:data_selection_more_budgets} shows that selecting the dataset with respect to \textsc{GeoSpOT} distances consistently outperform the random sampling baseline. For these additional budgets, selecting with \textsc{GeoSpOT} distances with an appropriate choice of $K$ either matches or outperforms the global baseline. 

For constrained dataset selection on the Geo-YFCC-Text dataset (\Cref{fig:text_data_selection}), increasing the number of selected domains ($K$) leads to improved performance for China, but yields little or no improvement for Brazil or the United States. Selecting with respect to \textsc{GeoSpOT}-BERT distances frequently outperforms other dataset selection methods, although for the United States at $K=2$, \textsc{GeoSpOT}-GeoCLIP and \textsc{GeoSpOT}-SatCLIP achieve higher accuracy.

\subsection{Applicability Maps}
\label{appx: aoa maps}

As mentioned in \Cref{appendix:dataset}, the Geo-YFCC-Image and Geo-YFCC-Text datasets differ in samples due to additional data cleaning for the text dataset. As a result, the corresponding \textsc{GeoSpOT} distances differ slightly between the two datasets. In the plotting of the applicability maps, we use the Geo-YFCC-Text \textsc{GeoSpOT} distances to plot all maps except for the ResNet50 map, for which we use the Geo-YFCC-Image dataset.

\begin{figure*}[!p]
\centering
\includegraphics[width=0.95\textwidth]{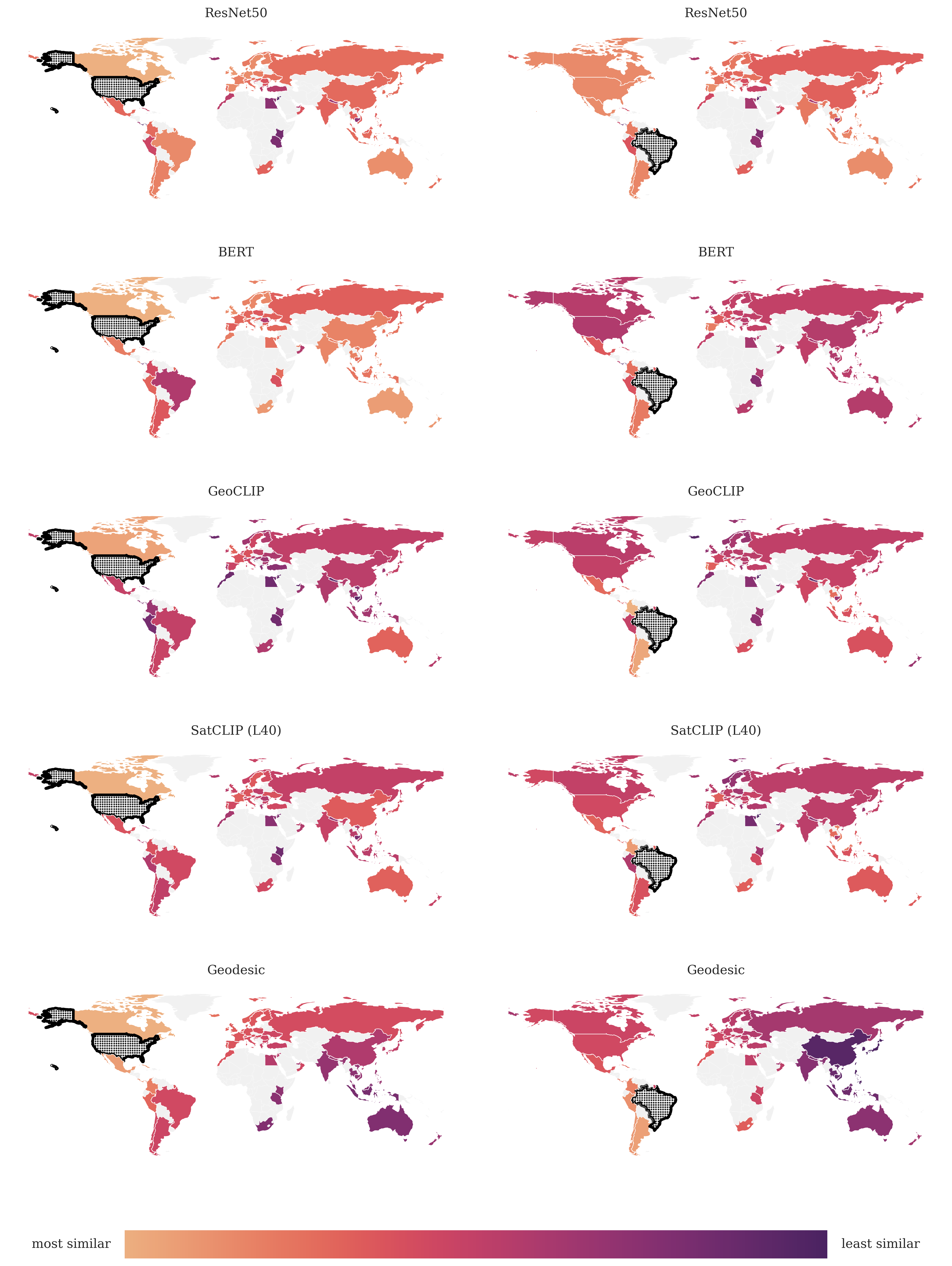}
\caption{Applicability maps for the \textbf{United States} (left) and \textbf{Brazil} (right).}
\label{fig:usa_brazil_maps}
\end{figure*}

\begin{figure*}[!p]
\centering
\includegraphics[width=0.95\textwidth]{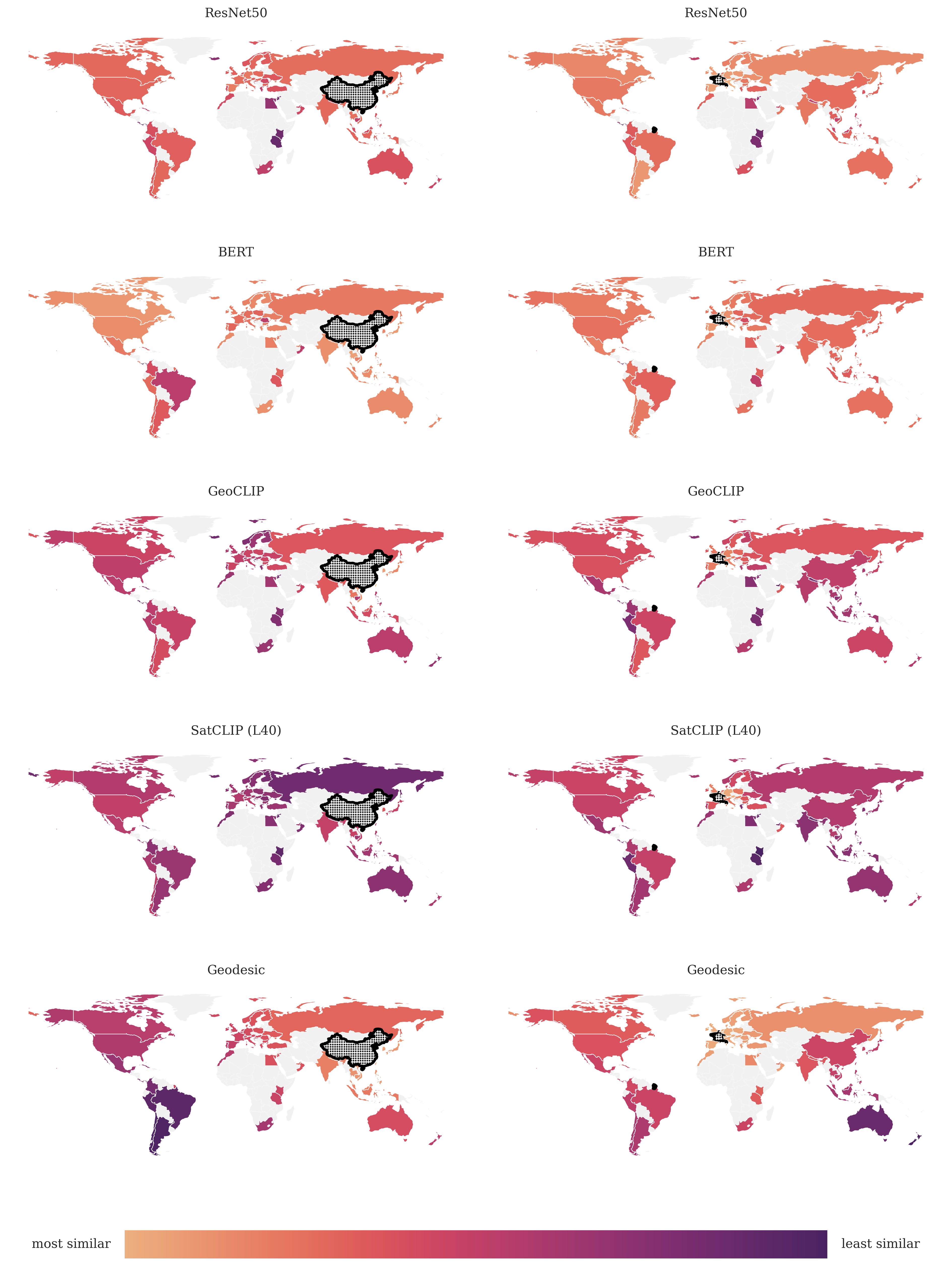}
\caption{Applicability maps for \textbf{China} (left) and \textbf{France} (right).}
\label{fig:china_france_maps}
\end{figure*}

We show additional applicability maps in \Cref{fig:usa_brazil_maps} and \Cref{fig:china_france_maps}. Because the maps vary substantially across countries, we render each set of maps as a larger figure to ensure maps can be viewed at sufficient resolution. Specifically, we show applicability maps for the United States, Brazil, China, and France across all \textsc{GeoSpOT} distances. As previously mentioned, we normalize the color scale for each embedding type.

\clearpage

\clearpage

\end{document}